\pdfoutput=1

\documentclass[10pt]{article}

\usepackage[utf8]{inputenc}
\usepackage[T1]{fontenc}
\usepackage{lmodern}          
\usepackage{microtype}

\usepackage[letterpaper,margin=1in]{geometry}

\usepackage{amsmath,amssymb,amsthm,mathtools}
\usepackage{bm}             

\usepackage{enumitem}

\usepackage{graphicx}
\graphicspath{{figures/}}     
\usepackage{booktabs}        
\usepackage{multirow}
\usepackage{subcaption}   
\usepackage{xcolor}
\usepackage[table]{xcolor}
\usepackage{parskip}
\usepackage{algorithm}
\usepackage{algpseudocode}
\usepackage{array}

\usepackage[numbers,sort&compress]{natbib}

\usepackage{fancyhdr}
\definecolor{stdgray}{gray}{0.55}

\fancypagestyle{plain}{\fancyhf{}\fancyfoot[C]{\thepage}}

\usepackage{tabularx}
\usepackage{authblk}

\usepackage[colorlinks=true,linkcolor=blue,citecolor=blue,urlcolor=blue]{hyperref}
\usepackage[capitalize,noabbrev]{cleveref}   
\crefname{algorithm}{Algorithm}{Algorithms}

\theoremstyle{definition}

\theoremstyle{remark}

\definecolor{darkblue}{RGB}{0, 0, 139}
\definecolor{gainpos}{RGB}{20,120,70}
\definecolor{gainneg}{RGB}{180,40,40}

\newcommand{\R}{\mathbb{R}}

\newcommand{\method}[1]{\textsc{#1}} 
\newcommand{\dataset}[1]{\texttt{#1}} 
\newcommand{\pmstd}[2]{#1{\scriptsize$\pm$#2}}

\DeclareMathOperator*{\argmin}{arg\,min}

\DeclareMathOperator{\sm}{sim}

\title{\textbf{FFM-CP: Cross-Backbone Fusion of Vision-Language Foundation Models for Few-Shot Computational Pathology}}

\author[1, 2]{Anh-Tien Nguyen}
\author[7]{Trung Dang\thanks{Second contribution author}}
\author[6]{ Nghiem Tuong Diep \protect\footnotemark[1]}
\author[6, 9]{Bui Ngoc Han Nguyen}
\author[12]{Tan-Ha Mai}
\author[2]{Miriam Cindy Maurer}
\author[10]{Phuong Hoa Nguyen}
\author[11]{Thi Thuy Uyen Nguyen}
\author[9]{Youngjun Park}
\author[5]{Daniel Sonntag}
\author[3, 4, 5]{Duy Minh Ho Nguyen}
\author[1]{ Anne-Christin Hauschild\thanks{Corresponding author: \texttt{anne-christin.hauschild@uni-giessen.de}}}

\affil[1]{Institute for Predictive Deep Learning for Medicine and Healthcare, Giessen University, Germany}
\affil[2]{Department of Medical Informatics, University Medicine Gottingen, Germany}
\affil[3]{Max Planck Research School for Intelligent Systems (IMPRS-IS), Germany}
\affil[4]{University of Stuttgart, Germany}
\affil[5]{German Research Center for Artificial Intelligence (DFKI), Germany}
\affil[6]{Department of machine learning, Mohamed bin Zayed University of Artificial Intelligence, UAE}
\affil[7]{Faculty of Information Technology and Electrical Engineering, University of Oulu, Finland}
\affil[8]{Carl von Ossietzky University of Oldenburg, Oldenburg, Germany}
\affil[9]{Max Planck Institute for Biology of Ageing, Cologne, Germany}
\affil[10]{Faculty of Basic Medicine and Pharmacy, VNU University of Medicine and Pharmacy, Vietnam}
\affil[11]{Department of Histology, Embryology, Pathology and Forensic Medicine, University of Medicine and Pharmacy, Hue University, Vietnam}
\affil[12]{Department of Computer Science and Information Engineering, National Taiwan University
, Taiwan}
\date{}
\hypersetup{
  pdftitle={FFM-CP: Cross-Backbone Fusion of Vision–Language Foundation Models for Few-Shot Computational Pathology},
  pdfauthor={First Author, Second Author, Third Author},
}

\begin{document}

\maketitle

\vspace{-2mm}

\begin{abstract}
\noindent
Pathology vision-language foundation models vary in performance across diseases and tasks, with no single model consistently performing best. The high cost of expert pathology annotation can also limit the labeled data available for task-specific adaptation. Combining complementary pretrained representations is a potential approach to these limitations, yet learning an effective fusion from few labeled examples remains challenging. We introduce \texttt{Few-shot Fusion Foundation Models of Computational Pathology} (\method{FFM-CP}), which is a framework that combines multiple pathology vision-language models in the few-shot learning setting. The framework first aligns heterogeneous representations using a closed-form Orthogonal Procrustes transformation estimated from corresponding support images. This alignment preserves within-model feature geometry without training an additional alignment network. Within the aligned space, a unified graph enables information exchange across backbones by jointly refining support-image features as well as visual and textual class prototypes. These refined representations support complementary text-prototype and case-retrieval branches that capture semantic class knowledge and within-class visual variation, respectively. Each branch learns to combine predictions from all ordered backbone pairs, allowing queries encoded by one model to draw on evidence represented by another. We evaluate three backbone combinations on six histopathology datasets at $4$, $8$, and $16$ shots per class. \method{FFM-CP} achieves higher mean macro-F1 than the strongest individually adapted member of each fused set in $50$ of $54$ comparisons. These findings suggest that combining complementary pretrained representations can improve histopathological classification when annotations are limited.
\end{abstract}


\newpage

\section{Introduction}
\label{sec:intro}
Histopathology images are central to cancer diagnosis, grading, and
subtyping, but developing computational models for these tasks traditionally requires large collections of images annotated by expert pathologists~\citep{campanella2019clinical,wahab2022semantic}. Such annotations are expensive, and they are particularly difficult to obtain for rare entities and institution-specific diagnostic problems. Pathology vision-language foundation models (VLMs) offer a promising alternative. By pretraining visual and text encoders on large image-text collections, these models learn reusable representations that can support zero-shot prediction and adaptation from only a small labeled support set
~\citep{ikezogwo2023quilt,zhang2025multimodal,lu2024visual,xiang2025vision,zhou2026knowledge}.

The growing number of pathology VLMs, however, creates a new model-selection problem. Their pretraining images, text sources, objectives, architectures, and embedding dimensions differ substantially. Consequently, each model captures a different subset of diagnostically useful morphology and terminology. Recent benchmarks confirm that model rankings vary across tissues and tasks, with no single foundation model consistently dominating the others \citep{bareja2026benchmark}. Selecting one backbone in advance therefore commits the downstream classifier to that model's representational strengths and blind spots while potentially discarding complementary evidence available in other models.

Existing studies address few-shot adaptation and multi-model fusion for computational pathology as separate problems. Few-shot methods tune prompts or lightweight adapters for a single pathology VLM ~\citep{he2026boosting, nguyen2025mgpath, shi2024vila, han2025mscpt, guo2025focus}, while the other methods combine independent pretrained vision and text encoders ~\citep{shalamflow,jiang2026exploring}. Conversely, multi-foundation-model approaches fuse visual features, slide representations, or prediction logits
~\citep{dang2024mfmf,luo2025ensemble,yang2025fusion}. These methods are generally designed for downstream cohorts containing hundreds or thousands of labeled samples, or they rely on a separate large-scale distillation stage. This leaves an important setting unresolved: how can several intact pathology VLMs be combined when the only downstream datasets consist of $K$ labeled tiles per class?


This setting presents three connected challenges. First, independent pathology-pretrained VLMs produce heterogeneous feature spaces. Their dimensions may differ, and even embeddings of the same dimension need not be directly comparable. Meaningful cross-model interaction therefore requires alignment estimated from a small support set. Second, the models should exchange class-specific evidence during adaptation. Aggregating predictions from independently adapted models does not allow support-image features or class prototypes to benefit from evidence represented by other backbones. Joint refinement would enable visual examples and textual class knowledge from different models to inform one another before prediction. Third, fusion should exploit both semantic class information and the visual variation among individual support examples. Moreover, the usefulness of these complementary sources can vary across backbones and tasks. Together, these challenges motivate a framework that aligns, jointly refines, and combines representations from multiple pathology VLMs in the few-shot learning setting.

We introduce \method{FFM-CP}, a support-set-driven framework for cross-backbone fusion of pathology VLMs. \method{FFM-CP} first uses a closed-form Orthogonal Procrustes transformation~\citep{schonemann1966generalized,lample2018word}, estimated from corresponding support images, to map all model representations into a shared reference space while preserving their internal geometry. It then constructs a unified heterogeneous graph containing support images, visual class prototypes, and text prototypes from every model. Seven relation types propagate visual and semantic evidence within and across backbones. Finally, cross-backbone pair fusion evaluates all $M^2$ ordered query-evidence combinations in two complementary branches: (1) a text-prototype branch transfers semantic class information, (2) a case-retrieval branch compares each query with its most relevant labeled examples and preserves intra-class variation. The foundation-model encoders remain frozen, and all task-specific components are learned using the few-shot support set alone.
Our contributions are threefold:
\begin{itemize}[itemsep=0pt, topsep=2pt, leftmargin=15pt, parsep=3pt]
    \item We formulate support-set-only fusion of heterogeneous pathology VLMs. To our knowledge, \method{FFM-CP} presents the first framework that adapts multiple pathology vision-language backbones in the few-shot learning setting.
    \item We combine parameter-free cross-backbone alignment, a unified multi-relation knowledge graph, and dual-branch fusion over all ordered model pairs, thereby exploiting complementary visual, textual, and case-based evidence.
    \item We evaluate five pathology VLMs and three fused model sets on six public datasets spanning four organs under $4$-, $8$-, and $16$-shot protocols. Across the extensive experiments, \method{FFM-CP} outperforms the best individual model within the corresponding fusion set in 50 cases, demonstrating the benefits of cross-backbone fusion.
\end{itemize}

\section{Related Work}
\label{sec:related}
\paragraph{Pathology vision-language foundation models.}
Contrastive image-text pretraining~\citep{radford2021learning} has been transferred to histopathology by a growing family of foundation models, whose training corpora range from medical Twitter threads~\citep{huanTwg2023visual}, narration transcribed from educational videos~\citep{ikezogwo2023quilt} to figure-caption pairs from the open-access literature~\citep{zhang2025multimodal}, in-house clinical collections~\citep{lu2024visual,xiang2025vision}, and released corpora restructured under a disease ontology~\citep{zhou2026knowledge}. Benchmarks agree that no member of this family dominates: across $31$ models and $41$ tasks, the top of the ranking changes with the tissue~\citep{bareja2026benchmark}.

\paragraph{Few-shot adaptation of vision-language models for histopathology images.}
Few-shot adaptation is particularly important in histopathology because assembling labeled whole-slide image datasets requires substantial clinical resources and expert effort. For frozen VLMs, \method{CoOp}~\citep{zhou2022conditional} learns prompt context vectors, while \method{CLIP-Adapter}~\citep{gao2024clip} adapts features through a lightweight bottleneck. \method{Tip-Adapter}~\citep{zhang2021tip} uses a training-free support-set cache, and \method{Graph Adapter}~\citep{li2023graphadapter} refines textual representations through a dual knowledge graph.

For WSI analysis, \method{ViLa-MIL}~\citep{shi2024vila} uses LLM-generated prompts at two magnification scales with $16$ labeled slides per class. \method{MSCPT}~\citep{han2025mscpt} learns multi-scale, context-aware prompts, while \method{FOCUS}~\citep{guo2025focus} uses language prompts to select diagnostically relevant patches. \method{MGPATH}~\citep{nguyen2025mgpath} couples the \method{Prov-GigaPath}~\citep{xu2024whole} vision encoder with the \method{PLIP}~\citep{huanTwg2023visual} text encoder through multi-granular prompting. \method{PathPT}~\citep{he2026boosting} applies prompt tuning to rare cancer subtyping with $1$--$10$ WSIs per subtype.
Related to our alignment stage, \method{FSF}~\citep{shalamflow} aligns independently pretrained visual and textual encoders using a closed-form orthogonal Procrustes transformation and regularizes adaptation with a flow-matching prior. These methods adapt a single backbone or a predefined vision-text pair. \method{FFM-CP} jointly refines visual and textual representations across $M$ frozen VLMs, enabling information exchange during adaptation, and combines predictions from all $M^2$ ordered backbone pairs.

\paragraph{Combining multiple foundation models in pathology.}
MFMF~\citep{dang2024mfmf} combines features from multiple patch encoders within a multiple-instance learning network for whole-slide classification. ELF~\citep{luo2025ensemble} integrates five tile-level foundation models through slide-encoder pretraining on $53{,}699$ WSIs from $11$ datasets. It combines contrastive alignment with weakly supervised objectives and uses fixed slide representations for downstream linear probing. FuseCPath~\citep{yang2025fusion} combines patch- and slide-level foundation models across scales. At the prediction level, LogitProd~\citep{huang2026plug} learns sample-adaptive weights over frozen experts' logits without aligning their feature spaces. Its training procedure first fits task-specific expert heads, then trains a fusion gate on a separate labeled subset. Meanwhile, $FM^2$~\citep{yu2025fm2} distills general and pathology foundation models into a single student by disentangling their consensus and divergence, followed by zero- and few-shot evaluation.

Our work focuses on learning task-specific fusion directly from a few-shot support set. \method{FFM-CP} retains each model's frozen visual and textual encoders and learns cross-model representation refinement and prediction fusion from the $C \cdot K$ labeled support tiles of a $C$-class, $K$-shot task. This approach requires no separate large-scale fusion pretraining or distillation stage.
\section{Method}
\label{sec:method}

\subsection{Preliminaries}
\label{sec:prelim}

\textbf{Vision-language foundation models for pathology.}
Contrastive image-text pretraining has been extended to histopathology~\citep{huanTwg2023visual,ikezogwo2023quilt,lu2024visual,zhou2026knowledge,chen2024towards,xiang2025vision,zhang2025multimodal}. These models typically comprise a visual encoder $E_v$ and a text encoder $E_t$, which map H\&E tiles and descriptions to representations $z_v$ and $z_t$, respectively. Pretraining encourages high cosine similarity between paired image and text representations. In our framework, both encoders remain frozen during downstream adaptation.

Despite this shared structure, their pretraining data, objectives, and architectures differ. Training corpora include medical social media posts~\citep{huanTwg2023visual}, transcribed educational videos~\citep{ikezogwo2023quilt}, article figure-caption pairs~\citep{zhang2025multimodal}, and non-public clinical cohorts~\citep{lu2024visual,xiang2025vision}. Most models directly align paired tiles and captions. MUSK~\citep{xiang2025vision} first applies unified masked modeling to unpaired data comprising $50$ million tiles and $10^{9}$ pathology-text tokens, then aligns the modalities. Text encoders range from generic CLIP-style Transformers to biomedical models informed by disease ontology~\citep{zhou2026knowledge}. Visual architectures and embedding dimensions also vary. These differences may contribute to task-dependent variation in model performance.


\textbf{Complementary strengths of pathology foundation models.}
A recent benchmark of 32 foundation models across 41 computational pathology tasks shows that no single model dominates consistently; instead, performance rankings vary substantially across tissues and tasks~\citep{bareja2026benchmark}. Different models perform best on brain and breast tumor classification, bladder and prostate subtyping, and lung-related tasks, suggesting that their learned representations capture different diagnostically useful signals.

This variation is also consistent with the diversity of morphological cues used across cancer types. Breast carcinoma grading considers tubule formation, nuclear pleomorphism, and mitotic activity~\citep{elston1991pathological}, whereas Gleason/ISUP grading of prostate cancer emphasizes glandular architecture~\citep{epstein20162014}. Lung adenocarcinoma subtyping, in contrast, depends largely on the predominant tissue growth pattern~\citep{travis2011international}. These differences do not fully explain why model rankings change within a given tissue, but they highlight that pathology tasks rely on distinct visual structures and scales. Together, this empirical and domain-level evidence motivates combining multiple pathology foundation models to exploit their complementary strengths.


\textbf{Few-shot multi-model setting.} We are given $M$ frozen pathology FMs indexed by $m\in \{1, \ldots,M\}$, each with its own encoders $E_v^m,E_t^m$, its own temperature $\tau_m$, and its own embedding dimension $d_m$. These dimensions can differ across the $M$ models. The support set $\mathcal{S}=\{(\mathbf{x}_i,y_i)\}_{i=1}^{N}$ contains $K$ labeled images from each of $C$ classes, giving $N=C \cdot K$ images in total. Here, $\mathbf{x}_i$ denotes the $i^{th}$ support image, and $y_i\in \{1,\ldots,C\}$ is its class label. We use $c \in \{1,\ldots,C\}$ to index the classes and $t_c$ to denote the text description of class $c$. For model $m$, the frozen encoders extract the visual and text features
\begin{equation}
z_{v,i}^{m}=E_v^{m}(\mathbf{x}_i)\in\R^{d_m},
\qquad
z_{t,c}^{m}=E_t^{m}(t_c)\in\R^{d_m},
\qquad
\bar z_{v,c}^{m}=\frac{1}{K}\sum_{i:\,y_i=c}z_{v,i}^{m},
\label{eq:extract}
\end{equation}
where $z_{v,i}^{m}$ is the vision features of the $i^{th}$-sample, $\bar z_{v,c}^{m}$ is the class-level visual prototype of class $c$, and $z_{t,c}^{m}$ is the textual features of the class description $t_c$. We collected the support features, visual prototypes, and text features as row matrices: We collect the feature vectors as rows:
\begin{equation}
Z_v^m=\bigl[(z_{v,i}^m)^\top\bigr]_{i=1}^{N},
\quad
\bar Z_v^m=\bigl[(\bar z_{v,c}^m)^\top\bigr]_{c=1}^{C},
\quad
Z_t^m=\bigl[(z_{t,c}^m)^\top\bigr]_{c=1}^{C}.
\label{eq:stack}
\end{equation}
Here, $Z_v^m\in\R^{N\times d_m}$ and $\bar Z_v^m,Z_t^m\in\R^{C\times d_m}$. A query tile $\mathbf{x}$ is encoded as $z_v^{m}(\mathbf{x})=E_v^{m}(\mathbf{x})$. A cross-model similarity such as $\sm(z_{v,i}^{m},z_{t,c}^{n})$ is dimensionally invalid when $d_m\ne d_n$ and is not geometrically meaningful. Even when their dimensions match, independently pretrained models may capture different visual semantic features. The same image or class description can therefore be represented differently across models, so direct cross-model similarities may not reliably reflect semantically corresponding features. This incompatibility is the first challenge addressed by our method.


\subsection{The proposed method}
\label{sec:approach-main}

\begin{figure}[H]
    \centering
    \vspace{-5pt}
    \includegraphics[width=0.99\linewidth]{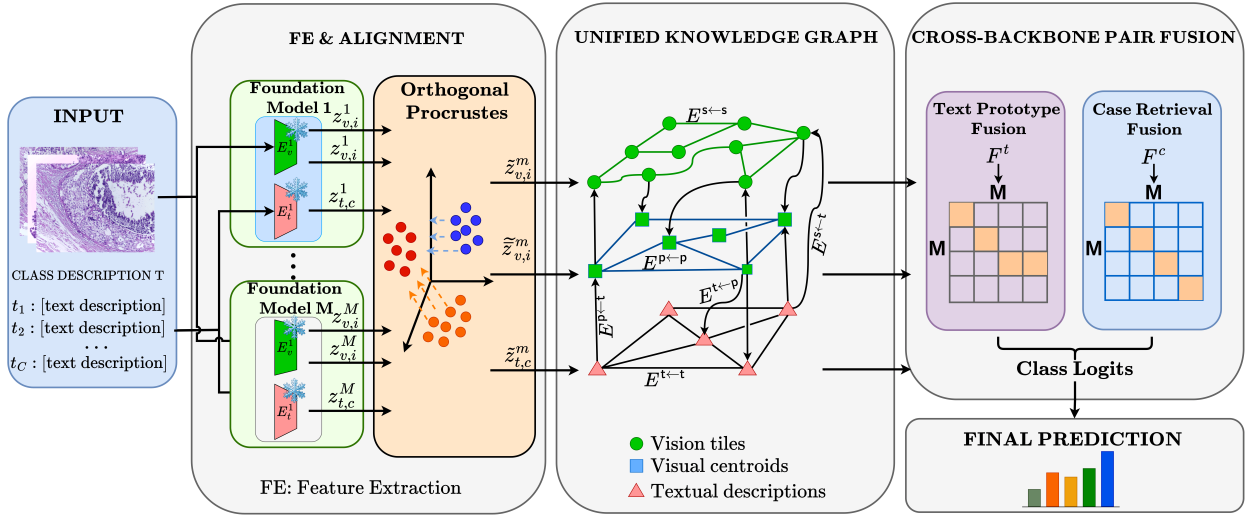}
    \caption{\textbf{Overview of FFM-CP}. Frozen pathology foundation models encode support tiles, class descriptions, and the query tile. Cross-backbone alignment projects their heterogeneous representations into a shared reference space, where a unified graph jointly refines support-image, visual-prototype, and text-prototype nodes through seven relation types. Finally, cross-backbone pair fusion aggregates all ordered query--evidence backbone pairs through separate text and case streams.}
    \label{fig:architecture}
    \vspace{-10pt}
\end{figure}

\method{FFM-CP} constructs a single heterogeneous graph $\mathcal{G}=(\mathcal{V},\mathcal{E})$ with three node types: \emph{(i)} support-image nodes carrying $z_{v,i}^{m}$, \emph{(ii)} text-prototype nodes carrying $z_{t,c}^{m}$, and \emph{(iii)} visual-prototype nodes carrying $\bar z_{v,c}^{m}$. The graph therefore contains $M \cdot N$ support-image nodes, $M \cdot C$ text-prototype nodes, and $M \cdot C$ visual-prototype nodes. Its edges may join nodes derived from different models, allowing a class description encoded by one model to be refined using visual evidence extracted by another. The full pipeline is illustrated in Figure~\ref{fig:architecture}.

\textbf{Cross-backbone alignment.}
As discussed in Section~\ref{sec:prelim}, pathology FMs may differ in embedding dimension and coordinate system.
Their features therefore require alignment before computing cross-backbone similarities in the unified graph.
\method{FFM-CP} uses Orthogonal Procrustes (OP) to map the features of each model into a shared reference space.
The semi-orthogonality constraint preserves norms and inner products, retaining the feature relationships learned during pretraining. OP admits a closed-form solution based on features extracted from the same support tiles.
This enables alignment using only the few-shot support set, without additional alignment data or iterative optimization. 


Let $r$ denote a reference model with the largest embedding dimension; thus, $d_r$ is the dimension of model $r$, and let $d=d_r$. For each model $m\ne r$, we obtain a semi-orthogonal map into the reference space by solving an Orthogonal Procrustes problem~\citep{schonemann1966generalized}:
\vspace{0.05in}
\begin{equation}
\mathbf{Q}_m=\argmin_{\mathbf{Q}\in\R^{d\times d_m},\;\mathbf{Q}^{\top}\mathbf{Q}=\mathbf{I}_{d_m}}
\bigl\|Z_v^{r}-Z_v^{m}\mathbf{Q}^{\top}\bigr\|_F^{2}
\;=\;\mathbf{U}_m\mathbf{V}_m^{\top},
\qquad
{Z_v^{r}}^{\top}Z_v^{m}=\mathbf{U}_m\bm{\Sigma}_m\mathbf{V}_m^{\top},
\label{eq:op}
\end{equation}
where $\mathbf{U}_m\bm{\Sigma}_m\mathbf{V}_m^{\top}$ denotes the thin singular value decomposition (SVD) of the cross-covariance. Here, $\mathbf{U}_m\in\R^{d\times d_m}$ and $\bm{\Sigma}_m,\mathbf{V}_m\in\R^{d_m\times d_m}$. We set $\mathbf{Q}_r=\mathbf{I}_{d}$. For any feature $z$ from model $m$, its aligned representation is $\tilde z=\mathbf{Q}_mz$. Applying this map to the support features and visual and textual class prototypes yields $\tilde Z_v^m$, $\tilde{\bar Z}_v^m$, and $\tilde Z_t^m$. The map estimated from support images also applies to textual features because both encoders share an embedding space within each model.

The condition $\mathbf{Q}_m^\top\mathbf{Q}_m=\mathbf{I}_{d_m}$ preserves feature norms and within-model cosine similarities. All aligned representations lie in $\R^d$, with cross-model alignment estimated from corresponding support images.

\vspace{0.05in}
\textbf{Unified knowledge graph.}
Our graph construction builds on \method{GraphAdapter}~\citep{li2023graphadapter}, which refines textual prompt features using task-specific structural knowledge encoded through inter-class relationships. We extend this idea by constructing a unified graph that jointly refines support-image features and visual and textual class prototypes across the $M$ pathology VLMs.

This structural modeling is particularly relevant to histopathology, where diagnostic categories are inherently related rather than independent. Tumor classifications often follow hierarchical structures~\citep{tan2020world}, while grading systems are ordinal, such that neighboring grades differ primarily in degree~\citep{elston1991pathological}. Consequently, adjacent or morphologically similar categories can share visual characteristics and are often difficult to distinguish in practice~\citep{brancati2022bracs}.

To explicitly capture these relationships, we construct a unified graph $\mathcal{G}=(\mathcal{V},\mathcal{E})$ that connects representations both within and across the $M$ backbones. Its node set contains three disjoint types,
$\mathcal{V}=\mathcal{V}^{\mathsf{t}}\cup\mathcal{V}^{\mathsf{p}}\cup\mathcal{V}^{\mathsf{s}}$,
corresponding to text prototypes, visual prototypes, and support images, respectively. We stack their aligned features model by model:
\begin{equation}
\mathbf{X}^{\mathsf{t}}
=\bigl[\tilde Z_t^1;\dots;\tilde Z_t^M\bigr],
\quad
\mathbf{X}^{\mathsf{p}}
=\bigl[\tilde{\bar Z}_v^1;\dots;\tilde{\bar Z}_v^M\bigr],
\quad
\mathbf{X}^{\mathsf{s}}
=\bigl[\tilde Z_v^1;\dots;\tilde Z_v^M\bigr].
\label{eq:nodes}
\end{equation}
Here, $\mathbf{X}^{\mathsf{t}},\mathbf{X}^{\mathsf{p}} \in\R^{MC\times d}$ and $\mathbf{X}^{\mathsf{s}}\in\R^{MN\times d}$. Within each model, image-text similarity provides class probabilities. For a tile $\mathbf{x}$ with representation $z_v^m=E_v^m(\mathbf{x})$, model $m$ computes
\begin{equation}
P_m(y=c\mid \mathbf{x})
=\frac{\exp\bigl(\tau_m\sm(z_v^m,z_{t,c}^m)\bigr)}
{\sum_{c'=1}^{C}
\exp\bigl(\tau_m\sm(z_v^m,z_{t,c'}^m)\bigr)}.
\label{eq:clip}
\end{equation}
The OP maps align the models using corresponding support images, allowing similarity comparisons across backbones. We use these aligned representations to construct the graph edges.

Let $a,b\in\{\mathsf{t},\mathsf{p},\mathsf{s}\}$ denote node types. The edge set $\mathcal{E}^{a\leftarrow b}\subseteq\mathcal{E}$ contains edges from nodes of type $b$ to nodes of type $a$. The graph contains six relations derived from feature similarities:
\begin{equation*}
\underbrace{
\mathcal{E}^{\mathsf{s}\leftarrow\mathsf{t}},\;
\mathcal{E}^{\mathsf{s}\leftarrow\mathsf{s}}
}_{\text{into support images}}
\qquad
\underbrace{
\mathcal{E}^{\mathsf{p}\leftarrow\mathsf{t}},\;
\mathcal{E}^{\mathsf{p}\leftarrow\mathsf{p}}
}_{\text{into visual prototypes}}
\qquad
\underbrace{
\mathcal{E}^{\mathsf{t}\leftarrow\mathsf{p}},\;
\mathcal{E}^{\mathsf{t}\leftarrow\mathsf{t}}
}_{\text{into text prototypes}}.
\end{equation*}


For each node type $a$, let $n_a=|\mathcal{V}^a|$ and let $\mathbf{X}_i^a$ denote the $i^{th}$ row of $\mathbf{X}^a$. Its mean feature is $\bm{\mu}^a=\frac{1}{n_a}\sum_{i=1}^{n_a}\mathbf{X}_i^a$. The weighted adjacency matrix $\mathbf{A}^{a\leftarrow b}\in\R^{n_a\times n_b}$ has receiving nodes in its rows and sending nodes in its columns. For the six feature-derived relations, we compute
\begin{equation}
    \bigl[\mathbf{A}^{a\leftarrow b}\bigr]_{ij}
    =\sm\bigl(
    \mathbf{X}_i^a-\bm{\mu}^a,\;
    \mathbf{X}_j^b-\bm{\mu}^b
    \bigr).
\label{eq:edges}
\end{equation}
For same-type relations ($a=b$), we set the diagonal entries
to zero to remove self-loops.

The seventh relation, $\mathcal{E}^{\mathsf{p}\leftarrow\mathsf{s}}$, uses the support labels. Each visual prototype links to the $K$ support-image nodes of the same class and backbone. Its binary adjacency matrix is
\begin{equation}
    \bigl[\mathbf{A}^{\mathsf{p}\leftarrow\mathsf{s}}\bigr]_{(m,c),(n,i)}
    =\mathbf{1}[m=n]\,\mathbf{1}[y_i=c].
    \label{eq:membership}
\end{equation}
Here, $(m,c)$ indexes the visual prototype of class $c$ from model $m$, and $(n,i)$ indexes support image $i$ represented by model $n$.

\vspace{0.05in}
\textbf{Graph learning across backbones.}
We sequentially refine support-image features, visual prototypes, and text prototypes. Each stage uses the latest available source features, and a superscript $*$ denotes refined features:
\vspace{0.05in}
\begin{equation}
\begin{aligned}
\mathbf{X}^{\mathsf{s},*}
&= \mathbf{X}^{\mathsf{s}}
+ w_1\,g_{\mathsf{s}}\bigl(
\mathbf{X}^{\mathsf{s}},\mathbf{X}^{\mathsf{t}},
\mathbf{A}^{\mathsf{s}\leftarrow\mathsf{t}}\bigr)
+ w_2\,g_{\mathsf{s}}\bigl(
\mathbf{X}^{\mathsf{s}},\mathbf{X}^{\mathsf{s}},
\mathbf{A}^{\mathsf{s}\leftarrow\mathsf{s}}\bigr),\\
\mathbf{X}^{\mathsf{p},*}
&= \mathbf{X}^{\mathsf{p}}
+ w_3\,g_{\mathsf{p}}\bigl(
\mathbf{X}^{\mathsf{p}},\mathbf{X}^{\mathsf{s},*},
\mathbf{A}^{\mathsf{p}\leftarrow\mathsf{s}}\bigr)
+ w_4\,g_{\mathsf{p}}\bigl(
\mathbf{X}^{\mathsf{p}},\mathbf{X}^{\mathsf{t}},
\mathbf{A}^{\mathsf{p}\leftarrow\mathsf{t}}\bigr)
+ w_5\,g_{\mathsf{p}}\bigl(
\mathbf{X}^{\mathsf{p}},\mathbf{X}^{\mathsf{p}},
\mathbf{A}^{\mathsf{p}\leftarrow\mathsf{p}}\bigr),\\
\mathbf{X}^{\mathsf{t},*}
&= \mathbf{X}^{\mathsf{t}}
+ w_6\,g_{\mathsf{t}}\bigl(
\mathbf{X}^{\mathsf{t}},\mathbf{X}^{\mathsf{p},*},
\mathbf{A}^{\mathsf{t}\leftarrow\mathsf{p},*}\bigr)
+ w_7\,g_{\mathsf{t}}\bigl(
\mathbf{X}^{\mathsf{t}},\mathbf{X}^{\mathsf{t}},
\mathbf{A}^{\mathsf{t}\leftarrow\mathsf{t}}\bigr).
\end{aligned}
\label{eq:cascade}
\end{equation}
Here, $\{w_i\}_{i=1}^{7}$ are learned relation scalars, and $g_{\mathsf{s}}$, $g_{\mathsf{p}}$, and $g_{\mathsf{t}}$ are destination-specific message-passing operators. Support-image nodes first receive messages from text prototypes and other support-image nodes. Visual prototypes then incorporate the refined support features, and text prototypes incorporate the refined visual prototypes. The feature-derived relations exchange information within and across backbones.

The adjacency $\mathbf{A}^{\mathsf{t}\leftarrow\mathsf{p},*}$ is recomputed using Equation~\ref{eq:edges} with $\mathbf{X}^{\mathsf{t}}$ and $\mathbf{X}^{\mathsf{p},*}$, including the updated mean of the visual prototypes. The remaining feature-derived adjacency matrices are computed from the initial aligned features and kept fixed. The membership adjacency $\mathbf{A}^{\mathsf{p}\leftarrow\mathsf{s}}$ is fixed by the support labels.

For destination and source feature matrices $\mathbf{X}\in\R^{n_X\times d}$ and $\mathbf{Y}\in\R^{n_Y\times d}$, respectively, let $\mathbf{A}\in\R^{n_X\times n_Y}$ be their weighted adjacency. For destination type $a\in\{\mathsf{s},\mathsf{p},\mathsf{t}\}$, we define
\begin{equation}
g_a(\mathbf{X},\mathbf{Y},\mathbf{A})
=\sigma\Bigl(
\mathbf{D}_X^{-1}\mathbf{X}\mathbf{W}_a^{\mathrm{self}}
+\widehat{\mathbf{A}}\mathbf{Y}\mathbf{W}_a^{\mathrm{msg}}
+\mathbf{b}_a
\Bigr).
\label{eq:gcn}
\end{equation}
The matrices $\mathbf{W}_a^{\mathrm{self}},\mathbf{W}_a^{\mathrm{msg}} \in\R^{d\times d}$ and bias $\mathbf{b}_a\in\R^d$ are learned separately for each destination type. They are shared across relations with the same destination type. The bias is added to each row, and $\sigma$ denotes an element-wise activation.

The normalized adjacency is
$\widehat{\mathbf{A}}
=\mathbf{D}_X^{-1/2}\mathbf{A}\mathbf{D}_Y^{-1/2}$.
The diagonal degree matrices are defined for each supplied adjacency by
\begin{equation*}
\mathbf{D}_X
=\operatorname{diag}\left(
1+\sum_{j=1}^{n_Y}|A_{ij}|
\right),
\qquad
\mathbf{D}_Y
=\operatorname{diag}\left(
1+\sum_{i=1}^{n_X}|A_{ij}|
\right).
\end{equation*}
Absolute edge weights accommodate signed similarities, while the added one ensures positive degrees.

\vspace{0.05in}
\textbf{Training objective.}
For support examples $(\mathbf{x}_i,y_i)$, we optimize $\mathcal{L}_{\mathrm{final}} =\mathcal{L}_{\mathrm{fuse}} + \eta \mathcal{L}_{\mathrm{diag}}$ where
\begin{equation}
\begin{aligned}
\mathcal{L}_{\mathrm{fuse}}
&=\frac{1}{N}\sum_{i=1}^{N}\!\Bigl[
\mathrm{CE}\bigl(\bm{\ell}^{\mathsf{t}}(\mathbf{x}_i),y_i\bigr)
+\lambda\,\mathrm{CE}\bigl(\bm{\ell}^{\mathsf{c}}(\mathbf{x}_i),y_i\bigr)\Bigr],\\
\mathcal{L}_{\mathrm{diag}}
&=\frac{1}{MN}\sum_{i=1}^{N}\sum_{m=1}^{M}\!\Bigl[
\mathrm{CE}\bigl(\bm{Z}^{\mathsf{t}}_{m,m}(\mathbf{x}_i),y_i\bigr)
+\lambda\,\mathrm{CE}\bigl(\bm{Z}^{\mathsf{c}}_{m,m}(\mathbf{x}_i),y_i\bigr)\Bigr].
\end{aligned}
\label{eq:loss}
\end{equation}
Here, $\mathrm{CE}$ denotes cross-entropy loss. The coefficient $\lambda$ weights the case-stream loss in both terms. The coefficient $\eta$ controls the contribution of the auxiliary diagonal loss. The diagonal term preserves each same-model classifier. We optimize the parameters of the message-passing operators $g_s$, $g_p$, and $g_t$, the relation weights $\{w_j\}_{j=1}^{7}$, and the pair-fusion matrices $\mathbf{F}^{\mathsf{t}}$ and $\mathbf{F}^{\mathsf{c}}$. All visual and textual encoders and Procrustes maps remain fixed. At inference, we apply softmax to each stream's fused logits and average the resulting class probabilities:
\begin{equation}
P(y=c\mid \mathbf{x})=\frac{1}{2}\Bigl[
\mathrm{softmax}\!\left(\bm{\ell}^{\mathsf{t}}(\mathbf{x})\right)_c+
\mathrm{softmax}\!\bigl(\bm{\ell}^{\mathsf{c}}(\mathbf{x})\bigr)_c
\Bigr].
\label{eq:prediction}
\end{equation}
\section{Experiments}
\label{sec:experiments}

\subsection{Experimental Setup}
\label{sec:experimental-setup}

\textbf{Datasets.}
\label{sec:datasets}
We evaluated the proposed method on six public histopathology datasets: \textbf{LungHist700}~\citep{diosdado2024lunghist700} ($C=7$ classes), \textbf{HeidelbergSkin}~\citep{kriegsmann2022deep} ($C=16$ classes), \textbf{Kather2016}~\citep{kather2016multi} ($C=8$ classes), \textbf{LubLung} ~\citep{rkaczkowska2022deep} ($C=9$ classes), \textbf{BRACS}~\citep{brancati2022bracs} ($C=7$ classes), \textbf{BACH}~\citep{aresta2019bach} ($C=4$ classes). The datasets span four organs, including lung, skin, colorectum, and breast. Moreover, they cover tissue-type recognition (Kather2016, LubLung, HeidelbergSkin) and diagnostic subtyping (LungHist700, BRACS, BACH), with sizes ranging from 400 images to 129{,}364 tiles and class counts from 4 to 16. Detailed descriptions are included in Appendix ~\ref{app:datasets}. 

Every dataset was divided into disjoint training, validation, and test proportions. Given a shot count $K$, we draw $K$ examples per class from the training partition, and $\min(K, 4)$ examples per class from the validation partition, following standard practice for few-shot adaptation of vision-language models ~\citep{li2023graphadapter}. For BRACS, we used the officially released test set as the test partition without further modification. Each experiment was repeated five times, and we provided the mean and standard deviation of these runs to ensure robustness and reproducibility.

\textbf{Implementation details.} 
In this study, we use the released pretrained weights of all foundation models and keep their encoders frozen throughout training. For each category, we average the text embeddings of $[P]$ prompt templates constructed from its diagnostic description $t_c$ (Appendix~\ref{app:add_imp_details}). The reference model $r$ is the member with the largest embedding dimension: \method{KEEP}~\citep{zhou2026knowledge} ($d=768$) for \method{KQP} and \method{KPB}, and \method{MUSK}~\citep{xiang2025vision} ($d=2048$) for \method{MQP}. The Procrustes maps in Equation ~\ref{eq:op} are computed once from the $C \cdot K$ support tiles. The loss weights are set to $\lambda=0.5$ and $\eta=0.5$ arcoss the datasets.

\textbf{Baseline methods.}
Following the criterion in Sec.~\ref{sec:prelim}, we evaluate each fused set against its strongest member. Each candidate model: \method{KEEP}~\citep{zhou2026knowledge}, \method{MUSK}~\citep{xiang2025vision}, \method{QuiltNet}~\citep{ikezogwo2023quilt}, \method{PLIP}~\citep{huanTwg2023visual}, and \method{BioMedCLIP}~\citep{zhang2025multimodal}, is adapted independently using the proposed pipeline with $M=1$. For each dataset and shot count, the fusion gain $\Delta$ is measured relative to the best-performing member of the corresponding fused set. We additionally evaluate a late-fusion ensemble in which each model compares query-tile features only with its own text prototypes and support-tile features. This ensemble is reported as variant~(a) of the ablation study in Sec.~\ref{sebsec:abla}.

\subsection{Results}
\label{subsec:results}
Tables~\ref{tab:f1_k16}, \ref{tab:f1_k8}, and~\ref{tab:f1_k4} report macro-F1 across six histopathology datasets at $16$, $8$, and $4$ shots, respectively. Results are presented as the mean and standard deviation over five random seeds. We evaluate three fused sets: KEEP$+$QuiltNet$+$PLIP (KQP), KEEP$+$PLIP$+$BiomedCLIP (KPB), and MUSK$+$QuiltNet$+$PLIP (MQP). All gains are expressed in percentage points relative to the strongest individual member of the corresponding fused set on each dataset.


\textbf{Fusion improves performance consistently.}
Across six datasets, three shot settings, and three fused sets, \method{FFM-CP} outperforms the strongest constituent in $50$ of $54$ comparisons, including all cases at $8$ and $16$ shots. Average gains increase from $2.35$ points at $4$ shots to $3.73$ points at $16$ shots, suggesting that cross-backbone fusion remains effective across few-shot regimes and benefits from a more informative support set.


\textbf{Fusion helps even when constituent models differ substantially in strength.}
\method{FFM-CP} can improve upon the strongest backbone even when the remaining models perform considerably worse on their own. For example, on LungHist700 at 16 shots, fusing KEEP, PLIP, and BiomedCLIP improves macro-F1 from $78.68\%$ for KEEP alone to $84.25\%$. This suggests that lower-performing models can still provide complementary evidence that strengthens the fused prediction.

\begin{table}[!t]
\centering
\vspace{-0.2in}
\caption{Macro F1 (\%) at $K=16$ shots on all six histopathology datasets. Mean $\pm$ standard deviation over 5 random seeds. \textbf{Bold} = best, \underline{underlined} = second best in each column. $\Delta$ indicates the gain over the strongest individual member of that same fused set, selected per column. Accuracy and AUC are in Appendix~\ref{app:vlmfull}.}
\vspace{0.05in}
\label{tab:f1_k16}
\setlength{\tabcolsep}{4pt}
\renewcommand{\arraystretch}{1.05}
\resizebox{\textwidth}{!}{%
\begin{tabular}{lcccccc|c}
\toprule
Method & LungHist700 & HeidelbergSkin & Kather2016 & LubLung & BRACS & BACH & Avg. \\
\midrule
\multicolumn{8}{c}{\textit{Vision--language foundation models (GCN adapter)}} \\
\midrule
KEEP & 78.68{\tiny$\,\pm$2.14} & 90.64{\tiny$\,\pm$0.99} & 87.28{\tiny$\,\pm$1.10} & 81.44{\tiny$\,\pm$1.28} & 52.11{\tiny$\,\pm$2.74} & 89.32{\tiny$\,\pm$1.37} & 79.91 \\
MUSK & 70.75{\tiny$\,\pm$2.68} & 83.27{\tiny$\,\pm$0.40} & 87.66{\tiny$\,\pm$0.21} & 75.73{\tiny$\,\pm$2.58} & 50.55{\tiny$\,\pm$3.78} & 84.00{\tiny$\,\pm$1.12} & 75.33 \\
QuiltNet & 66.76{\tiny$\,\pm$2.90} & 75.50{\tiny$\,\pm$0.28} & 86.64{\tiny$\,\pm$0.51} & 74.64{\tiny$\,\pm$1.16} & 45.78{\tiny$\,\pm$3.12} & 72.93{\tiny$\,\pm$4.48} & 70.37 \\
PLIP & 67.19{\tiny$\,\pm$2.26} & 75.51{\tiny$\,\pm$0.61} & 86.34{\tiny$\,\pm$0.84} & 74.53{\tiny$\,\pm$2.85} & 43.03{\tiny$\,\pm$2.54} & 72.60{\tiny$\,\pm$3.71} & 69.87 \\
BiomedCLIP & 61.72{\tiny$\,\pm$1.05} & 69.66{\tiny$\,\pm$1.36} & 81.40{\tiny$\,\pm$2.10} & 68.96{\tiny$\,\pm$2.62} & 49.42{\tiny$\,\pm$2.60} & 68.49{\tiny$\,\pm$4.30} & 66.61 \\
\midrule
\multicolumn{8}{c}{\textit{Fused VLMs (ours: OP alignment + combined graph + GCN)}} \\
\midrule
KEEP$+$QuiltNet$+$PLIP & \underline{84.11}{\tiny$\,\pm$1.63} & \textbf{92.68}{\tiny$\,\pm$0.32} & \textbf{91.66}{\tiny$\,\pm$1.56} & \textbf{86.86}{\tiny$\,\pm$0.99} & \underline{52.77}{\tiny$\,\pm$1.96} & \underline{92.18}{\tiny$\,\pm$2.20} & \underline{83.37} \\
\quad $\Delta$ & \textcolor{darkblue}{\small $+5.43$} & \textcolor{darkblue}{\small $+2.04$} & \textcolor{darkblue}{\small $+4.38$} & \textcolor{darkblue}{\small $+5.42$} & \textcolor{darkblue}{\small $+0.66$} & \textcolor{darkblue}{\small $+2.86$} & \textcolor{darkblue}{\small $+3.47$} \\
\cmidrule(lr){1-8}
KEEP$+$PLIP$+$BiomedCLIP & \textbf{84.25}{\tiny$\,\pm$1.38} & \underline{92.58}{\tiny$\,\pm$0.65} & \underline{91.63}{\tiny$\,\pm$1.52} & \underline{86.42}{\tiny$\,\pm$0.88} & \textbf{54.53}{\tiny$\,\pm$3.04} & \textbf{92.38}{\tiny$\,\pm$1.86} & \textbf{83.63} \\
\quad $\Delta$ & \textcolor{darkblue}{\small $+5.57$} & \textcolor{darkblue}{\small $+1.94$} & \textcolor{darkblue}{\small $+4.35$} & \textcolor{darkblue}{\small $+4.98$} & \textcolor{darkblue}{\small $+2.42$} & \textcolor{darkblue}{\small $+3.06$} & \textcolor{darkblue}{\small $+3.72$} \\
\cmidrule(lr){1-8}
MUSK$+$QuiltNet$+$PLIP & 78.67{\tiny$\,\pm$2.10} & 87.25{\tiny$\,\pm$0.70} & 91.19{\tiny$\,\pm$0.76} & 81.73{\tiny$\,\pm$1.00} & 52.00{\tiny$\,\pm$2.68} & 85.19{\tiny$\,\pm$0.74} & 79.34 \\
\quad $\Delta$ & \textcolor{darkblue}{\small $+7.92$} & \textcolor{darkblue}{\small $+3.98$} & \textcolor{darkblue}{\small $+3.53$} & \textcolor{darkblue}{\small $+6.00$} & \textcolor{darkblue}{\small $+1.45$} & \textcolor{darkblue}{\small $+1.19$} & \textcolor{darkblue}{\small $+4.01$} \\
\bottomrule
\end{tabular}}
\vspace{-10pt}
\end{table}

\begin{table}[!t]
\centering
\caption{Macro F1 (\%) at $K=8$ shots on all six histopathology datasets. Mean $\pm$ standard deviation over 5 random seeds. \textbf{Bold} = best, \underline{underlined} = second best in each column. Accuracy and AUC are in Appendix~\ref{app:vlmfull}.}
\vspace{0.05in}
\label{tab:f1_k8}
\setlength{\tabcolsep}{4pt}
\renewcommand{\arraystretch}{1.05}
\resizebox{\textwidth}{!}{%
\begin{tabular}{lcccccc|c}
\toprule
Method & LungHist700 & HeidelbergSkin & Kather2016 & LubLung & BRACS & BACH & Avg. \\
\midrule
\multicolumn{8}{c}{\textit{Vision-language foundation models (GCN adapter)}} \\
\midrule
KEEP & 75.62{\tiny$\,\pm$6.87} & 87.90{\tiny$\,\pm$0.61} & 86.83{\tiny$\,\pm$1.08} & 77.53{\tiny$\,\pm$1.20} & 50.96{\tiny$\,\pm$2.34} & 89.35{\tiny$\,\pm$3.14} & 78.03 \\
MUSK & 67.60{\tiny$\,\pm$3.87} & 79.86{\tiny$\,\pm$3.38} & 85.17{\tiny$\,\pm$2.03} & 72.00{\tiny$\,\pm$2.19} & 44.99{\tiny$\,\pm$0.49} & 81.47{\tiny$\,\pm$4.08} & 71.85 \\
QuiltNet & 57.93{\tiny$\,\pm$9.33} & 70.89{\tiny$\,\pm$1.12} & 83.56{\tiny$\,\pm$1.41} & 71.76{\tiny$\,\pm$0.55} & 42.31{\tiny$\,\pm$2.24} & 67.82{\tiny$\,\pm$4.94} & 65.71 \\
PLIP & 60.28{\tiny$\,\pm$4.97} & 71.52{\tiny$\,\pm$1.32} & 82.66{\tiny$\,\pm$1.74} & 70.84{\tiny$\,\pm$1.19} & 40.61{\tiny$\,\pm$1.85} & 66.91{\tiny$\,\pm$9.57} & 65.47 \\
BiomedCLIP & 56.28{\tiny$\,\pm$3.78} & 64.68{\tiny$\,\pm$1.38} & 77.77{\tiny$\,\pm$2.36} & 63.30{\tiny$\,\pm$0.84} & 43.44{\tiny$\,\pm$2.33} & 70.52{\tiny$\,\pm$1.70} & 62.66 \\
\midrule
\multicolumn{8}{c}{\textit{Fused VLMs (ours: OP alignment + combined graph + GCN)}} \\
\midrule
KEEP$+$QuiltNet$+$PLIP & \textbf{77.40}{\tiny$\,\pm$2.95} & \underline{90.74}{\tiny$\,\pm$1.12} & \textbf{90.87}{\tiny$\,\pm$1.11} & \textbf{83.48}{\tiny$\,\pm$0.83} & \textbf{53.23}{\tiny$\,\pm$3.31} & \underline{89.53}{\tiny$\,\pm$2.29} & \textbf{80.88} \\
\quad $\Delta$ & \textcolor{darkblue}{\small $+1.78$} & \textcolor{darkblue}{\small $+2.84$} & \textcolor{darkblue}{\small $+4.04$} & \textcolor{darkblue}{\small $+5.95$} & \textcolor{darkblue}{\small $+2.27$} & \textcolor{darkblue}{\small $+0.18$} & \textcolor{darkblue}{\small $+2.84$} \\
\cmidrule(lr){1-8}
KEEP$+$PLIP$+$BiomedCLIP & \underline{76.98}{\tiny$\,\pm$2.67} & \textbf{91.48}{\tiny$\,\pm$1.23} & \underline{90.53}{\tiny$\,\pm$1.02} & \underline{82.36}{\tiny$\,\pm$1.30} & \underline{52.30}{\tiny$\,\pm$2.19} & \textbf{89.78}{\tiny$\,\pm$2.77} & \underline{80.57} \\
\quad $\Delta$ & \textcolor{darkblue}{\small $+1.36$} & \textcolor{darkblue}{\small $+3.58$} & \textcolor{darkblue}{\small $+3.7$} & \textcolor{darkblue}{\small $+4.83$} & \textcolor{darkblue}{\small $+1.34$} & \textcolor{darkblue}{\small $+0.43$} & \textcolor{darkblue}{\small $+2.54$} \\
\cmidrule(lr){1-8}
MUSK$+$QuiltNet$+$PLIP & 72.21{\tiny$\,\pm$2.87} & 84.25{\tiny$\,\pm$0.34} & 88.82{\tiny$\,\pm$0.97} & 78.59{\tiny$\,\pm$1.14} & 48.47{\tiny$\,\pm$1.16} & 82.60{\tiny$\,\pm$4.35} & 75.82 \\
\quad $\Delta$ & \textcolor{darkblue}{\small $+4.61$} & \textcolor{darkblue}{\small $+4.39$} & \textcolor{darkblue}{\small $+3.65$} & \textcolor{darkblue}{\small $+6.59$} & \textcolor{darkblue}{\small $+3.48$} & \textcolor{darkblue}{\small $+1.13$} & \textcolor{darkblue}{\small $+3.98$} \\
\bottomrule
\end{tabular}}
\vspace{-10pt}
\end{table}

\subsection{Ablation Studies}
\label{sebsec:abla}

We examine the contributions of cross-backbone prediction pairs and feature alignment using $K=16$ support examples per class and $M=3$ backbones: KEEP, QuiltNet, and PLIP. Table~\ref{tab:ablation} compares the full model with three variants. The diagonal variant restricts both prediction streams to same-backbone pairs ($m=n$) while retaining the full graph. The other two variants replace Procrustes alignment with zero-padding or a two-layer MLP for each model. The MLPs are trained jointly with the graph.


\textbf{Contribution of cross-backbone prediction pairs.}
Including cross-backbone query--evidence pairs yields a modest average gain of $0.59$ macro-F1 points, with improvements on four of six datasets. The effect is therefore smaller and more task-dependent than that of representation alignment, suggesting that cross-model prediction transfer is helpful when backbones provide complementary evidence, but is not uniformly beneficial across datasets.

\begin{table}[!t]
\centering
\vspace{-0.25in}
\caption{Macro F1 (\%) at $K=4$ shots on all six histopathology datasets. Mean $\pm$ standard deviation over 5 random seeds. \textbf{Bold} = best, \underline{underlined} = second best in each column. Accuracy and AUC are in Appendix~\ref{app:vlmfull}.}
\vspace{0.05in}
\label{tab:f1_k4}
\setlength{\tabcolsep}{4pt}
\renewcommand{\arraystretch}{1.05}
\resizebox{\textwidth}{!}{%
\begin{tabular}{lcccccc|c}
\toprule
Method & LungHist700 & HeidelbergSkin & Kather2016 & LubLung & BRACS & BACH & Avg. \\
\midrule
\multicolumn{8}{c}{\textit{Vision--language foundation models (GCN adapter)}} \\
\midrule
KEEP & 65.33{\tiny$\,\pm$3.02} & 86.26{\tiny$\,\pm$0.83} & 76.40{\tiny$\,\pm$1.66} & 72.10{\tiny$\,\pm$4.99} & \underline{48.64}{\tiny$\,\pm$4.91} & \textbf{86.26}{\tiny$\,\pm$9.94} & 72.50 \\
MUSK & 60.52{\tiny$\,\pm$4.39} & 74.49{\tiny$\,\pm$2.75} & 79.44{\tiny$\,\pm$3.30} & 61.51{\tiny$\,\pm$3.18} & 38.67{\tiny$\,\pm$2.09} & 75.66{\tiny$\,\pm$6.89} & 65.05 \\
QuiltNet & 53.15{\tiny$\,\pm$1.89} & 63.89{\tiny$\,\pm$1.27} & 75.42{\tiny$\,\pm$3.93} & 62.44{\tiny$\,\pm$1.68} & 34.83{\tiny$\,\pm$2.90} & 52.17{\tiny$\,\pm$6.43} & 56.98 \\
PLIP & 50.94{\tiny$\,\pm$4.08} & 64.76{\tiny$\,\pm$0.57} & 76.34{\tiny$\,\pm$2.79} & 61.50{\tiny$\,\pm$1.50} & 32.35{\tiny$\,\pm$2.63} & 55.72{\tiny$\,\pm$7.00} & 56.94 \\
BiomedCLIP & 52.62{\tiny$\,\pm$7.02} & 58.00{\tiny$\,\pm$3.83} & 70.72{\tiny$\,\pm$5.14} & 55.96{\tiny$\,\pm$0.75} & 37.65{\tiny$\,\pm$2.81} & 62.04{\tiny$\,\pm$8.48} & 56.16 \\
\midrule
\multicolumn{8}{c}{\textit{Fused VLMs (ours: OP alignment + combined graph + GCN)}} \\
\midrule
KEEP$+$QuiltNet$+$PLIP & \textbf{66.76}{\tiny$\,\pm$6.63} & \underline{88.30}{\tiny$\,\pm$1.71} & 82.47{\tiny$\,\pm$3.62} & \textbf{76.08}{\tiny$\,\pm$4.62} & 48.33{\tiny$\,\pm$2.32} & 84.61{\tiny$\,\pm$3.57} & \underline{74.43} \\
\quad $\Delta$ & \textcolor{darkblue}{\small $+1.43$} & \textcolor{darkblue}{\small $+2.04$} & \textcolor{darkblue}{\small $+6.07$} & \textcolor{darkblue}{\small $+3.98$} & \textcolor{gainneg}{\small $-0.31$} & \textcolor{gainneg}{\small $-1.65$} & \textcolor{darkblue}{\small $+1.93$} \\
\cmidrule(lr){1-8}
KEEP$+$PLIP$+$BiomedCLIP & \underline{66.19}{\tiny$\,\pm$6.87} & \textbf{88.38}{\tiny$\,\pm$2.23} & \underline{82.69}{\tiny$\,\pm$3.48} & \underline{76.02}{\tiny$\,\pm$3.67} & \textbf{49.53}{\tiny$\,\pm$2.69} & \underline{85.94}{\tiny$\,\pm$2.78} & \textbf{74.79} \\
\quad $\Delta$ & \textcolor{darkblue}{\small $+0.86$} & \textcolor{darkblue}{\small $+2.12$} & \textcolor{darkblue}{\small $+6.29$} & \textcolor{darkblue}{\small $+3.92$} & \textcolor{darkblue}{\small $+0.89$} & \textcolor{gainneg}{\small $-0.32$} & \textcolor{darkblue}{\small $+2.29$} \\
\cmidrule(lr){1-8}
MUSK$+$QuiltNet$+$PLIP & 62.08{\tiny$\,\pm$4.37} & 77.87{\tiny$\,\pm$1.24} & \textbf{82.88}{\tiny$\,\pm$4.36} & 68.98{\tiny$\,\pm$2.28} & 42.22{\tiny$\,\pm$3.48} & 74.23{\tiny$\,\pm$5.80} & 68.04 \\
\quad $\Delta$ & \textcolor{darkblue}{\small $+1.56$} & \textcolor{darkblue}{\small $+3.38$} & \textcolor{darkblue}{\small $+3.44$} & \textcolor{darkblue}{\small $+6.54$} & \textcolor{darkblue}{\small $+3.55$} & \textcolor{gainneg}{\small $-1.43$} & \textcolor{darkblue}{\small $+2.84$} \\
\bottomrule
\end{tabular}}
\vspace{-5pt}
\end{table}

\vspace{-0.15in}
\begin{table}[!t]
\centering
\caption{Ablation of \method{FFM-CP} at $K=16$ shots using KEEP, QuiltNet, and PLIP.
(a) Restricts prediction fusion to diagonal pairs ($m=n$) in both streams while retaining the full graph.
(b) Replaces the Procrustes map in Equation~\ref{eq:op} with zero-padding to $\mathbb{R}^{d}$.
(c) Replaces Procrustes with a two-layer MLP per model, trained jointly with the graph. Results are macro-F1 (\%) scores.}
\vspace{0.05in}
\label{tab:ablation}
\footnotesize
\setlength{\tabcolsep}{3pt}
\resizebox{0.95\linewidth}{!}{%
\begin{tabular}{@{}l cccccc@{}}
\toprule
Configure & LungHist700 & HeidelbergSkin & Kather2016 & LubLung & BRACS & BACH \\
\midrule
\method{} \method{FFM-CP(full)} &
 \textbf{\pmstd{84.11}{1.63}} & \pmstd{92.68}{0.32} & \textbf{\pmstd{91.66}{1.56}} &
 \textbf{\pmstd{86.86}{0.99}} & \pmstd{52.77}{1.96} & \textbf{\pmstd{92.18}{2.20}}  \\
\midrule
(a) \method{FFM-CP(diagonal)} &
 \pmstd{82.24}{1.76} & \textbf{\pmstd{92.71}{0.31}} & \pmstd{90.61}{1.84} &
 \pmstd{86.77}{1.01} & \textbf{\pmstd{52.79}}{2.15} & \pmstd{91.58}{1.93} \\
(b) \method{zero-padding} &
 \pmstd{80.94}{1.16} & \pmstd{91.63}{0.20} & \pmstd{90.09}{1.14} &
 \pmstd{85.22}{1.01} & \pmstd{51.79}{2.59} & \pmstd{90.22}{2.35} \\
(c) \method{MLP}&
 \pmstd{79.73}{2.53} & \pmstd{90.60}{0.79} & \pmstd{90.34}{0.95} &
 \pmstd{85.30}{0.58} & \pmstd{51.03}{0.85} & \pmstd{90.54}{1.82} \\
\bottomrule
\end{tabular}}
\vspace{-0.2in}
\end{table}

\vspace{0.15in}
\textbf{Contribution of Procrustes alignment.}
Procrustes alignment consistently outperforms zero-padding and learned MLP alignment, improving mean macro-F1 by $1.73$ and $2.12$ points on average, respectively. While zero-padding only matches feature dimensions, Procrustes additionally aligns the backbones' coordinate systems while preserving their internal geometry. This suggests that, in the few-shot regime, preserving pretrained structure while establishing cross-model correspondence is more effective than learning a flexible alignment from limited supervision.
\section{Conclusion}
\label{sec:conclusion}

We introduced FFM-CP to study few-shot fusion of heterogeneous pathology foundation models. Our results show that even weaker individual backbones can contribute complementary diagnostic evidence when their representations are aligned and jointly refined. Ablations further suggest that preserving pretrained geometry through simple Procrustes alignment is more effective than learning a flexible alignment under limited supervision. Overall, heterogeneity across pathology foundation models can be an asset: combining complementary frozen representations can yield stronger few-shot predictions than relying on any single model.

\textbf{Limitations and Broader Impacts.}
Our evaluation is limited to six public classification datasets and a fixed set of pathology VLMs; broader validation on additional backbones, whole-slide tasks, and cross-institution cohorts remains future work. Considering all $M^2$ ordered backbone pairs also increases fusion cost as the model set grows. Moreover, Procrustes assumes a global linear orthogonal relation between feature spaces, which may not capture more complex cross-model discrepancies. Finally, because alignment and graph construction rely on the few-shot support set, performance may depend on how representative those examples are, especially in extremely low-shot settings. The framework may reduce reliance on task-specific annotations. However, potential biases inherited from the pretrained models and performance differences across patient groups require evaluation before clinical use.

\section*{Acknowledgements}
This work is supported in part by funds from the German Ministry of Education and Research (BMBF) under grant agreements \textit{No. 01D2208A} and \textit{No. 01KD2414A} (project FAIrPaCT). The authors gratefully acknowledge the computing time granted by the KISSKI project. The calculations for this research were conducted with computing resources under the project \textit{kisski-umg-fairpact-2}.  The authors also acknowledge the computing time granted by the Resource Allocation Board and provided on the supercomputer Emmy/Grete at NHR-Nord@Göttingen as part of the NHR infrastructure. The calculations for this research were conducted with computing resources under the project \textit{nim00014}. Anh-Tien Nguyen was a member of the Ph.D. program "Genome Science" - International Max Planck Research School. \\
We gratefully acknowledge support from the hessian.AI Service Center (funded by the Federal Ministry of Research, Technology and Space, BMFTR, grant no. 16IS22091) and the hessian.AI Innovation Lab (funded by the Hessian Ministry for Digital Strategy and Innovation, grant no. S-DIW04/0013/003).

\bibliographystyle{unsrtnat}   
\bibliography{references}

\newpage
\appendix
\definecolor{darkblue}{RGB}{0, 0, 139}
\definecolor{gainneg}{RGB}{180,40,40}
\definecolor{stdgray}{RGB}{130,130,130}
\clearpage
\appendix
\begin{center}{\Large\scshape Appendix}\end{center}
\vspace{0.5em}


\section{Datasets}
\label{app:datasets}

\paragraph{LungHist700.}~\citep{diosdado2024lunghist700} contains 691 high-resolution $1200 \times 1600$ lung images from 45 patients at Hospital Cl\'inico Universitario de
Valladolid in 2023 during routine diagnostics, captured at both $20\times$ and $40\times$ magnification. The tissues are categorized into three superclasses, namely adenocarcinoma (\texttt{aca}), squamous cell carcinoma (\texttt{scc}), and normal lung (\texttt{nor}). Each carcinoma is further graded as well, moderately or poorly differentiated. 

\paragraph{HeidelbergSkin.}~\citep{kriegsmann2022deep}. The dataset comprises 129{,}364 tiles of
$\approx 395 \times 395$~px annotated in QuPath from 386 cases. The dataset was drawn from the archives of the Institute of Pathology, Heidelberg University, the MVZ for Histology, Cytology and Molecular Diagnostics Trier, and the Institute for Dermatopathology Hannover. Diagnoses followed the WHO Classification of Skin Tumors. The patches were annotated for 16 categories: chondral tissue, dermis, elastosis, epidermis, hair follicle, skeletal muscle, necrosis, nerves, sebaceous glands, subcutis, eccrine glands, vessels, BCC, SqCC, naevi and melanoma.

\paragraph{Kather2016.}~\citep{kather2016multi}. The dataset is a colorectal cancer texture dataset collected from the pathology archive of the Institute of Pathology, University Medical Center Mannheim, Heidelberg UniversitEight tissue types were selected, namely tumor epithelium, simple stroma, complex stroma, immune cell conglomerates, debris, normal mucosal glands, adipose tissue, and background, with 625 tiles each, giving 5{,}000 images of $150 \times 150$~px. 

\paragraph{LubLung.} ~\citep{rkaczkowska2022deep}. The dataset comprises 23,199 tiles of $172 \times 172$~px derived from the formalin-fixed, paraffin-embedded (FFPE) tissue samples from 55 primary tumors of lung cancer at Medical University of Lublin, Poland. The tiles were annotated by an expert pathologist with nine labels: tumor, stroma, mixed, immune, vessel, bronchi, necrosis, lung, and background.

\paragraph{BRACS.}~\citep{brancati2022bracs} provides 4{,}539 regions of interest
extracted from 547 whole-slide images from 189 patients between 2019 and 2020 at the Department of Pathology at the National Cancer Institute - Scientific Institute for Research, Hospitalization and Healthcare (IRCCS) `Fondazione G. Pascale`. Each WSI and its ROIs were annotated by the consensus of three board-certified pathologists into three lesion types: benign, atypical, and malignant. Then, the tiles are subtyped into seven categories: Normal, Pathological Benign, Usual Ductal Hyperplasia, Flat Epithelial Atypia, Atypical Ductal Hyperplasia, Ductal Carcinoma In Situ, and Invasive Carcinoma. Evaluation uses the released test partition of 570 ROIs unchanged. In this study, few-shot sampling draws the $K$-shot support from the 3{,}657 training ROIs and a validation subset from the 312 validation ROIs. We used the released test set proportion for evaluating the performance of the training process.

\paragraph{BACH.}~\citep{aresta2019bach} is the ICIAR 2018 breast histology challenge set: 400 microscopy images of $2048 \times 1536$~px at $0.42$~\textmu m/px, balanced across normal, benign, \emph{in situ} carcinoma and invasive carcinoma at 100 images each. Its four balanced classes make it the coarsest task we evaluate.

The six public datasets used in this study cover a diverse range of organs, tissue types, image resolutions, and dataset sizes, providing a comprehensive benchmark for evaluating the generalizability of the proposed method. Table~\ref{tab:datasets} summarizes the key characteristics of these datasets, including their organ of origin, number of classes, and total number of samples.

\begin{table}[H]
\centering
\caption{The description of the six datasets used in the study.}
\label{tab:datasets}
\setlength{\tabcolsep}{5pt}
\renewcommand{\arraystretch}{1.15}
\begin{tabular}{llcc}
\toprule
Dataset & Organ & Classes & Samples  \\
\midrule
LungHist700~\citep{diosdado2024lunghist700}    & Lung       & 7  & 691              \\
HeidelbergSkin~\citep{kriegsmann2022deep} & Skin       & 16 & 129{,}364        \\
Kather2016~\citep{kather2016multi}     & Colorectum & 8  & 5{,}000          \\
LubLung~\citep{rkaczkowska2022deep}        & Lung       & 9 & 23{,}199  \\
BRACS~\citep{brancati2022bracs}          & Breast     & 7  & 4{,}539    \\
BACH  ~\citep{aresta2019bach}         & Breast     & 4  & 400            \\
\bottomrule
\end{tabular}
\end{table}

\section{Additional Implementation Details}
~\label{app:add_imp_details}
In this paper, we used the prompts listed in Table \ref{tab:prompt-templates} for training and testing.

\begin{table}[H]
\centering
\caption{The number of classes and the prompt templates used for each dataset.
The five prompt embeddings of a class are averaged into its textual feature $z_{t,c}^{m}$}
\label{tab:prompt-templates}
\footnotesize
\setlength{\tabcolsep}{4pt}
\begin{tabular}{@{}l c >{\raggedright\arraybackslash}p{3.55in}@{}}
\toprule
Datasets & \# Classes & Prompt Template \\
\midrule
LungHist700~\citep{diosdado2024lunghist700}       & 7  & ``an H\&E stained lung histopathology image showing [class].''$^{\dagger}$ \\
HeidelbergSkin~\citep{kriegsmann2022deep} & 15 & ``an H\&E stained skin histopathology image showing [class].''$^{\dagger}$ \\
Kather2016~\citep{kather2016multi}         & 8  & ``an H\&E stained colorectal histopathology image showing [class].''$^{\dagger}$ \\
LubLung~\citep{rkaczkowska2022deep}               & 9  & ``an H\&E stained lung histopathology image showing [class].''$^{\dagger}$ \\
BRACS~\citep{brancati2022bracs}                   & 7  & ``an H\&E stained breast histopathology image showing [class].''$^{\dagger}$ \\
BACH~\citep{aresta2019bach}                     & 4  & ``an H\&E stained breast histopathology image showing [class].''$^{\dagger}$ \\
\midrule
\multicolumn{3}{@{}p{6.0in}@{}}{$^{\dagger}$Each dataset uses an ensemble of five templates: the one listed and ``a [organ] histopathology image showing [class].'', ``a microscopic image of [organ] tissue showing [class].'', ``a pathology slide of [organ] showing [class].'' and ``[class] in [organ] histopathology, H\&E stain.'', with [organ] the organ word of the listed template (lung, skin, colorectal or breast).} \\
\bottomrule
\end{tabular}
\end{table}

\section{Vision-language backbones and their fusions}
\label{app:vlmfull}

Tables ~\ref{app:vlm_k16_a} and ~\ref{app:vlm_k16_b} complement macro-F1 results with accuracy and AUC at $K=16$ on \dataset{LungHist700}, \dataset{HeidelbergSkin}, \dataset{Kather2016}, \dataset{LubLung}, \dataset{BRACS}, and \dataset{BACH} in Table ~\ref{tab:f1_k16}. For each dataset and metric, $\Delta$ is calculated relative to the strongest constituent of the corresponding fused set. 

\textbf{\dataset{LungHist700}.}
\method{KPB} achieves the highest mean macro-F1 ($84.25\%$) and AUC ($98.15\%$), improving upon its strongest constituent by $5.57$ and $1.35$ percentage points, respectively. \method{KQP} yields similar scores ($84.11\%$ and $98.09\%$), and both configurations achieve $85.76\%$ accuracy. \method{MQP} produces the largest gains over its own constituents: $7.61$, $7.92$, and $2.31$ percentage points in accuracy, macro-F1, and AUC, respectively.

\textbf{\dataset{HeidelbergSkin}.}
On \dataset{HeidelbergSkin}, the strongest constituent AUC is already $99.09$-$99.48\%$, leaving limited room for absolute improvement. \method{KQP} achieves the highest mean accuracy ($93.28\%$), macro-F1 ($92.68\%$), and AUC ($99.79\%$), with gains of $2.26$, $2.04$, and $0.31$ percentage points over \method{KEEP}. \method{KPB} follows closely, differing by at most $0.10$ percentage points across the three metrics. \method{MQP} improves macro-F1 from MUSK's $83.27\%$ to $87.25\%$, although this remains below \method{KEEP} alone ($90.64\%$). Across the three fused sets, AUC increases by $0.30$-$0.40$ percentage points from already high constituent scores of $99.09$-$99.48\%$. These small absolute AUC gains accompany bigger improvements in accuracy ($2.22$-$3.29$ percentage points) and macro-F1 ($1.94$-$3.98$ percentage points).

\textbf{\dataset{Kather2016}.}
All three fused sets exceed $91\%$ in both accuracy and macro-F1, with gains of $3.44$-$4.37$ and $3.53$-$4.38$ percentage points, respectively. \method{KQP} achieves the highest mean accuracy ($91.65\%$), macro-F1 ($91.66\%$), and AUC ($99.27\%$). The fused configurations are closely grouped, with differences of at most $0.51$, $0.47$, and $0.10$ percentage points across the three metrics. AUC improves by $0.41$-$0.69$ percentage points over the strongest constituent for each fused set. Although MUSK is the strongest individual backbone on this dataset across all three metrics, \method{KQP} and \method{KPB} attain slightly higher fused scores than \method{MQP}. The ranking of individual backbones therefore does not directly determine the ranking of their fusions.

\begin{table}[H]
\centering
\caption{Accuracy, macro F1, and AUC (\%) at $K=16$ shots on \dataset{LungHist700}, \dataset{HeidelbergSkin}, \dataset{Kather2016}, for the vision-language backbones and the fused sets whose members are all vision-language models. Mean $\pm$ standard deviation over 5 random seeds. \textbf{Bold} = best, \underline{underlined} = second best in each column.}
\label{app:vlm_k16_a}
\setlength{\tabcolsep}{2.6pt}
\renewcommand{\arraystretch}{1.05}
\resizebox{\textwidth}{!}{%
\begin{tabular}{lccccccccc}
\toprule
Method & \multicolumn{3}{c}{LungHist700} & \multicolumn{3}{c}{HeidelbergSkin} & \multicolumn{3}{c}{Kather2016} \\
\cmidrule(lr){2-4}\cmidrule(lr){5-7}\cmidrule(lr){8-10}
& ACC & F1 & AUC & ACC & F1 & AUC & ACC & F1 & AUC \\
\midrule
\multicolumn{10}{l}{\textit{Vision--language foundation models (GCN adapter)}} \\
KEEP & 80.39{\tiny$\,\pm$1.80} & 78.68{\tiny$\,\pm$2.14} & 96.80{\tiny$\,\pm$0.51} & 91.02{\tiny$\,\pm$1.12} & 90.64{\tiny$\,\pm$0.99} & 99.48{\tiny$\,\pm$0.10} & 87.28{\tiny$\,\pm$1.13} & 87.28{\tiny$\,\pm$1.10} & 98.54{\tiny$\,\pm$0.21} \\
MUSK & 72.98{\tiny$\,\pm$2.59} & 70.75{\tiny$\,\pm$2.68} & 95.04{\tiny$\,\pm$0.77} & 85.64{\tiny$\,\pm$0.53} & 83.27{\tiny$\,\pm$0.40} & 99.09{\tiny$\,\pm$0.07} & 87.70{\tiny$\,\pm$0.21} & 87.66{\tiny$\,\pm$0.21} & 98.76{\tiny$\,\pm$0.10} \\
QuiltNet & 69.56{\tiny$\,\pm$2.01} & 66.76{\tiny$\,\pm$2.90} & 91.43{\tiny$\,\pm$1.21} & 79.42{\tiny$\,\pm$0.78} & 75.50{\tiny$\,\pm$0.28} & 98.01{\tiny$\,\pm$0.14} & 86.66{\tiny$\,\pm$0.48} & 86.64{\tiny$\,\pm$0.51} & 97.82{\tiny$\,\pm$0.17} \\
PLIP & 69.95{\tiny$\,\pm$2.71} & 67.19{\tiny$\,\pm$2.26} & 91.55{\tiny$\,\pm$1.38} & 79.90{\tiny$\,\pm$0.66} & 75.51{\tiny$\,\pm$0.61} & 98.23{\tiny$\,\pm$0.08} & 86.42{\tiny$\,\pm$0.78} & 86.34{\tiny$\,\pm$0.84} & 98.58{\tiny$\,\pm$0.13} \\
BiomedCLIP & 65.27{\tiny$\,\pm$1.56} & 61.72{\tiny$\,\pm$1.05} & 90.74{\tiny$\,\pm$1.20} & 73.96{\tiny$\,\pm$1.38} & 69.66{\tiny$\,\pm$1.36} & 96.60{\tiny$\,\pm$0.21} & 81.29{\tiny$\,\pm$2.03} & 81.40{\tiny$\,\pm$2.10} & 97.59{\tiny$\,\pm$0.33} \\
\midrule
\multicolumn{10}{l}{\textit{Fused sets (ours: OP alignment + combined graph + GCN)}} \\
KEEP$+$QuiltNet$+$PLIP & \underline{85.76}{\tiny$\,\pm$1.46} & \underline{84.11}{\tiny$\,\pm$1.63} & \underline{98.09}{\tiny$\,\pm$0.47} & \textbf{93.28}{\tiny$\,\pm$0.46} & \textbf{92.68}{\tiny$\,\pm$0.32} & \textbf{99.79}{\tiny$\,\pm$0.02} & \textbf{91.65}{\tiny$\,\pm$1.54} & \textbf{91.66}{\tiny$\,\pm$1.56} & \textbf{99.27}{\tiny$\,\pm$0.23} \\
\quad $\Delta$ & \textcolor{darkblue}{\small $+5.37$} & \textcolor{darkblue}{\small $+5.43$} & \textcolor{darkblue}{\small $+1.29$} & \textcolor{darkblue}{\small $+2.26$} & \textcolor{darkblue}{\small $+2.04$} & \textcolor{darkblue}{\small $+0.31$} & \textcolor{darkblue}{\small $+4.37$} & \textcolor{darkblue}{\small $+4.38$} & \textcolor{darkblue}{\small $+0.69$} \\
\cmidrule(lr){1-10}
KEEP$+$PLIP$+$BiomedCLIP & \textbf{85.76}{\tiny$\,\pm$1.32} & \textbf{84.25}{\tiny$\,\pm$1.38} & \textbf{98.15}{\tiny$\,\pm$0.34} & \underline{93.24}{\tiny$\,\pm$0.68} & \underline{92.58}{\tiny$\,\pm$0.65} & \underline{99.78}{\tiny$\,\pm$0.03} & \underline{91.61}{\tiny$\,\pm$1.51} & \underline{91.63}{\tiny$\,\pm$1.52} & \underline{99.25}{\tiny$\,\pm$0.15} \\
\quad $\Delta$ & \textcolor{darkblue}{\small $+5.37$} & \textcolor{darkblue}{\small $+5.57$} & \textcolor{darkblue}{\small $+1.35$} & \textcolor{darkblue}{\small $+2.22$} & \textcolor{darkblue}{\small $+1.94$} & \textcolor{darkblue}{\small $+0.30$} & \textcolor{darkblue}{\small $+4.33$} & \textcolor{darkblue}{\small $+4.35$} & \textcolor{darkblue}{\small $+0.67$} \\
\cmidrule(lr){1-10}
MUSK$+$QuiltNet$+$PLIP & 80.59{\tiny$\,\pm$1.80} & 78.67{\tiny$\,\pm$2.10} & 97.35{\tiny$\,\pm$0.39} & 88.93{\tiny$\,\pm$1.00} & 87.25{\tiny$\,\pm$0.70} & 99.49{\tiny$\,\pm$0.06} & 91.14{\tiny$\,\pm$0.74} & 91.19{\tiny$\,\pm$0.76} & 99.17{\tiny$\,\pm$0.11} \\
\quad $\Delta$ & \textcolor{darkblue}{\small $+7.61$} & \textcolor{darkblue}{\small $+7.92$} & \textcolor{darkblue}{\small $+2.31$} & \textcolor{darkblue}{\small $+3.29$} & \textcolor{darkblue}{\small $+3.98$} & \textcolor{darkblue}{\small $+0.4$} & \textcolor{darkblue}{\small $+3.44$} & \textcolor{darkblue}{\small $+3.53$} & \textcolor{darkblue}{\small $+0.41$} \\
\bottomrule
\end{tabular}}
\end{table}

\textbf{\dataset{LubLung}.}
\method{KQP} achieves the highest mean accuracy ($90.00\%$), macro-F1 ($86.86\%$), and AUC ($98.92\%$), improving upon \method{KEEP} by $4.38$, $5.42$, and $1.10$ percentage points, respectively. \method{KPB} follows closely, with differences of $0.25$, $0.44$, and $0.05$ percentage points across the three metrics. \method{MQP} achieves the largest macro-F1 gain over its constituents ($6.00$ percentage points). All three fused sets also show lower standard deviations than their strongest constituent across all three metrics, indicating less variation across the five random seeds.

\textbf{\dataset{BRACS}.}
\method{KPB} achieves the highest mean accuracy ($56.25\%$), macro-F1 ($54.53\%$), and AUC ($88.13\%$), with gains of $3.79$, $2.42$, and $2.57$ percentage points over \method{KEEP}. Its macro-F1 exceeds \method{KQP} by $1.76$ percentage points, the largest difference between these two configurations across the six datasets. \method{KQP} yields a smaller macro-F1 improvement of $0.66$ percentage points, alongside a $2.12$-percentage-point increase in AUC. \method{MQP} improves all three metrics over MUSK, but its macro-F1 ($52.00\%$) remains slightly below \method{KEEP} alone ($52.11\%$). Thus, all fused sets improve upon their own constituents, while \method{KPB} provides the largest gains across all three metrics on this dataset.

\begin{table}[H]
\centering
\caption{Accuracy, macro F1 and AUC (\%) at $K=16$ shots on \dataset{LubLung}, \dataset{BRACS}, \dataset{BACH}, for the vision-language backbones and the fused sets whose members are all vision-language models. Mean $\pm$ standard deviation over 5 random seeds. \textbf{Bold} = best, \underline{underlined} = second best in each column.}
\label{app:vlm_k16_b}
\setlength{\tabcolsep}{2.6pt}
\renewcommand{\arraystretch}{1.05}
\resizebox{\textwidth}{!}{%
\begin{tabular}{lccccccccc}
\toprule
Method & \multicolumn{3}{c}{LubLung} & \multicolumn{3}{c}{BRACS} & \multicolumn{3}{c}{BACH} \\
\cmidrule(lr){2-4}\cmidrule(lr){5-7}\cmidrule(lr){8-10}
& ACC & F1 & AUC & ACC & F1 & AUC & ACC & F1 & AUC \\
\midrule
\multicolumn{10}{l}{\textit{Vision--language foundation models (GCN adapter)}} \\
KEEP & 85.62{\tiny$\,\pm$1.54} & 81.44{\tiny$\,\pm$1.28} & 97.82{\tiny$\,\pm$0.34} & 52.46{\tiny$\,\pm$2.08} & 52.11{\tiny$\,\pm$2.74} & 85.56{\tiny$\,\pm$0.85} & 89.33{\tiny$\,\pm$1.43} & 89.32{\tiny$\,\pm$1.37} & 98.35{\tiny$\,\pm$0.42} \\
MUSK & 80.16{\tiny$\,\pm$2.83} & 75.73{\tiny$\,\pm$2.58} & 96.66{\tiny$\,\pm$0.36} & 50.49{\tiny$\,\pm$4.20} & 50.55{\tiny$\,\pm$3.78} & 84.53{\tiny$\,\pm$1.80} & 83.83{\tiny$\,\pm$1.08} & 84.00{\tiny$\,\pm$1.12} & 95.62{\tiny$\,\pm$0.26} \\
QuiltNet & 79.67{\tiny$\,\pm$1.25} & 74.64{\tiny$\,\pm$1.16} & 96.22{\tiny$\,\pm$0.32} & 46.25{\tiny$\,\pm$3.37} & 45.78{\tiny$\,\pm$3.12} & 79.71{\tiny$\,\pm$1.64} & 72.50{\tiny$\,\pm$4.59} & 72.93{\tiny$\,\pm$4.48} & 88.64{\tiny$\,\pm$3.58} \\
PLIP & 79.31{\tiny$\,\pm$2.10} & 74.53{\tiny$\,\pm$2.85} & 96.11{\tiny$\,\pm$0.48} & 43.16{\tiny$\,\pm$3.08} & 43.03{\tiny$\,\pm$2.54} & 78.91{\tiny$\,\pm$1.13} & 72.33{\tiny$\,\pm$3.75} & 72.60{\tiny$\,\pm$3.71} & 88.30{\tiny$\,\pm$1.06} \\
BiomedCLIP & 74.52{\tiny$\,\pm$3.21} & 68.96{\tiny$\,\pm$2.62} & 94.43{\tiny$\,\pm$0.76} & 49.51{\tiny$\,\pm$2.50} & 49.42{\tiny$\,\pm$2.60} & 82.17{\tiny$\,\pm$1.19} & 68.67{\tiny$\,\pm$4.21} & 68.49{\tiny$\,\pm$4.30} & 87.13{\tiny$\,\pm$1.77} \\
\midrule
\multicolumn{10}{l}{\textit{Fused sets (ours: OP alignment + combined graph + GCN)}} \\
KEEP$+$QuiltNet$+$PLIP & \textbf{90.00}{\tiny$\,\pm$0.76} & \textbf{86.86}{\tiny$\,\pm$0.99} & \textbf{98.92}{\tiny$\,\pm$0.10} & \underline{54.07}{\tiny$\,\pm$1.41} & \underline{52.77}{\tiny$\,\pm$1.96} & \underline{87.68}{\tiny$\,\pm$0.63} & \underline{92.17}{\tiny$\,\pm$2.19} & \underline{92.18}{\tiny$\,\pm$2.20} & \underline{99.33}{\tiny$\,\pm$0.14} \\
\quad $\Delta$ & \textcolor{darkblue}{\small $+4.38$} & \textcolor{darkblue}{\small $+5.42$} & \textcolor{darkblue}{\small $+1.1$} & \textcolor{darkblue}{\small $+1.61$} & \textcolor{darkblue}{\small $+0.66$} & \textcolor{darkblue}{\small $+2.12$} & \textcolor{darkblue}{\small $+2.84$} & \textcolor{darkblue}{\small $+2.86$} & \textcolor{darkblue}{\small $+0.98$} \\
\cmidrule(lr){1-10}
KEEP$+$PLIP$+$BiomedCLIP & \underline{89.75}{\tiny$\,\pm$0.70} & \underline{86.42}{\tiny$\,\pm$0.88} & \underline{98.87}{\tiny$\,\pm$0.08} & \textbf{56.25}{\tiny$\,\pm$2.65} & \textbf{54.53}{\tiny$\,\pm$3.04} & \textbf{88.13}{\tiny$\,\pm$0.69} & \textbf{92.33}{\tiny$\,\pm$1.85} & \textbf{92.38}{\tiny$\,\pm$1.86} & \textbf{99.37}{\tiny$\,\pm$0.15} \\
\quad $\Delta$ & \textcolor{darkblue}{\small $+4.13$} & \textcolor{darkblue}{\small $+4.98$} & \textcolor{darkblue}{\small $+1.05$} & \textcolor{darkblue}{\small $+3.79$} & \textcolor{darkblue}{\small $+2.42$} & \textcolor{darkblue}{\small $+2.57$} & \textcolor{darkblue}{\small $+3.00$} & \textcolor{darkblue}{\small $+3.06$} & \textcolor{darkblue}{\small $+1.02$} \\
\cmidrule(lr){1-10}
MUSK$+$QuiltNet$+$PLIP & 85.65{\tiny$\,\pm$0.84} & 81.73{\tiny$\,\pm$1.00} & 98.20{\tiny$\,\pm$0.17} & 53.02{\tiny$\,\pm$2.70} & 52.00{\tiny$\,\pm$2.68} & 86.46{\tiny$\,\pm$0.70} & 85.17{\tiny$\,\pm$0.68} & 85.19{\tiny$\,\pm$0.74} & 97.02{\tiny$\,\pm$1.10} \\
\quad $\Delta$ & \textcolor{darkblue}{\small $+5.49$} & \textcolor{darkblue}{\small $+6.00$} & \textcolor{darkblue}{\small $+1.54$} & \textcolor{darkblue}{\small $+2.53$} & \textcolor{darkblue}{\small $+1.45$} & \textcolor{darkblue}{\small $+1.93$} & \textcolor{darkblue}{\small $+1.34$} & \textcolor{darkblue}{\small $+1.19$} & \textcolor{darkblue}{\small $+1.40$} \\
\bottomrule
\end{tabular}}
\end{table}

\textbf{\dataset{BACH}.}
\method{KPB} achieves the highest mean accuracy ($92.33\%$), macro-F1 ($92.38\%$), and AUC ($99.37\%$), improving upon \method{KEEP} by $3.00$, $3.06$, and $1.02$ percentage points, respectively. \method{KQP} performs similarly, trailing by only $0.16$, $0.20$, and $0.04$ percentage points. \method{MQP} yields smaller accuracy and macro-F1 gains ($1.34$ and $1.19$ percentage points), despite achieving the largest AUC gain ($1.40$ percentage points).

Tables ~\ref{app:vlm_k8_a} and ~\ref{app:vlm_k8_b} report accuracy, macro-F1, and AUC at $K=8$ on \dataset{LungHist700}, \dataset{HeidelbergSkin}, and \dataset{Kather2016}, \dataset{LubLung}, \dataset{BRACS}, and  \dataset{BACH} in the setting of $K=8$ shots.

\textbf{\dataset{LungHist700}.}
\method{KQP} achieves the highest mean accuracy ($79.63\%$) and macro-F1 ($77.40\%$), improving upon \method{KEEP} by $1.09$ and $1.78$ percentage points, respectively. \method{KPB} attains a marginally higher AUC ($96.61\%$ versus $96.58\%$), while its macro-F1 is $76.98\%$. For these two fusions, macro-F1 gains over \method{KEEP} are $1.36$-$1.78$ percentage points, smaller than the $5.43$-$5.57$ gains at $K=16$. At eight shots, both also reduce the macro-F1 standard deviation from $6.87$ for \method{KEEP} to $2.95$ for \method{KQP} and $2.67$ for \method{KPB}. \method{MQP} achieves the largest macro-F1 gain over its constituents ($4.61$ percentage points), but its final score ($72.21\%$) remains below \method{KEEP} alone ($75.62\%$).

\textbf{\dataset{HeidelbergSkin}.}
\method{KPB} achieves the highest mean accuracy ($92.02\%$), macro-F1 ($91.48\%$), and AUC ($99.69\%$), with gains of $2.69$, $3.58$, and $0.51$ percentage points over \method{KEEP}. \method{KQP} follows with $90.74\%$ macro-F1 and $99.67\%$ AUC. The difference between these two configurations is therefore more pronounced in macro-F1 ($0.74$ percentage points) than in AUC ($0.02$ percentage points). \method{MQP} produces the largest macro-F1 gain over its constituents ($4.39$ percentage points), although its final score ($84.25\%$) remains below the two \method{KEEP}-based fusions. Despite these differences in final macro-F1, all three fused sets achieve similar AUC gains of $0.49$-$0.52$ percentage points.

\textbf{\dataset{Kather2016}.}
\method{KQP} achieves the highest mean accuracy ($90.84\%$), macro-F1 ($90.87\%$), and AUC ($99.10\%$), improving upon \method{KEEP} by $4.04$, $4.04$, and $0.72$ percentage points, respectively. \method{KPB} follows closely with $90.53\%$ macro-F1, corresponding to a gain of $3.70$ percentage points over \method{KEEP}. MQP reaches $88.82\%$ macro-F1, improving upon MUSK by $3.65$ percentage points and also exceeding \method{KEEP} alone ($86.83\%$). All three fused sets improve AUC by $0.50$-$0.72$ percentage points. Relative to $K=16$, macro-F1 is lower by $0.79$ percentage points for \method{KQP} and $1.10$ for \method{KPB}, compared with $2.37$ for \method{MQP}.

\begin{table}[H]
\centering
\caption{Accuracy, macro F1, and AUC (\%) at $K=8$ shots on \dataset{LungHist700}, \dataset{HeidelbergSkin}, \dataset{Kather2016}, for the vision-language backbones and the fused sets whose members are all vision-language models. Mean $\pm$ standard deviation over 5 random seeds. \textbf{Bold} = best, \underline{underlined} = second best in each column.}
\label{app:vlm_k8_a}
\setlength{\tabcolsep}{2.6pt}
\renewcommand{\arraystretch}{1.05}
\resizebox{\textwidth}{!}{%
\begin{tabular}{lccccccccc}
\toprule
Method & \multicolumn{3}{c}{LungHist700} & \multicolumn{3}{c}{HeidelbergSkin} & \multicolumn{3}{c}{Kather2016} \\
\cmidrule(lr){2-4}\cmidrule(lr){5-7}\cmidrule(lr){8-10}
& ACC & F1 & AUC & ACC & F1 & AUC & ACC & F1 & AUC \\
\midrule
\multicolumn{10}{l}{\textit{Vision-language foundation models (GCN adapter)}} \\
KEEP & 78.54{\tiny$\,\pm$5.63} & 75.62{\tiny$\,\pm$6.87} & 95.68{\tiny$\,\pm$0.98} & 89.33{\tiny$\,\pm$0.67} & 87.90{\tiny$\,\pm$0.61} & 99.18{\tiny$\,\pm$0.17} & 86.80{\tiny$\,\pm$0.83} & 86.83{\tiny$\,\pm$1.08} & 98.38{\tiny$\,\pm$0.12} \\
MUSK & 70.34{\tiny$\,\pm$3.39} & 67.60{\tiny$\,\pm$3.87} & 92.74{\tiny$\,\pm$0.35} & 83.19{\tiny$\,\pm$3.02} & 79.86{\tiny$\,\pm$3.38} & 98.60{\tiny$\,\pm$0.32} & 85.16{\tiny$\,\pm$1.80} & 85.17{\tiny$\,\pm$2.03} & 98.36{\tiny$\,\pm$0.44} \\
QuiltNet & 61.83{\tiny$\,\pm$8.05} & 57.93{\tiny$\,\pm$9.33} & 89.03{\tiny$\,\pm$2.52} & 76.07{\tiny$\,\pm$1.09} & 70.89{\tiny$\,\pm$1.12} & 96.80{\tiny$\,\pm$0.27} & 83.59{\tiny$\,\pm$1.55} & 83.56{\tiny$\,\pm$1.41} & 97.12{\tiny$\,\pm$0.42} \\
PLIP & 62.54{\tiny$\,\pm$4.49} & 60.28{\tiny$\,\pm$4.97} & 90.18{\tiny$\,\pm$1.22} & 76.68{\tiny$\,\pm$1.47} & 71.52{\tiny$\,\pm$1.32} & 97.53{\tiny$\,\pm$0.20} & 82.75{\tiny$\,\pm$1.62} & 82.66{\tiny$\,\pm$1.74} & 97.68{\tiny$\,\pm$0.54} \\
BiomedCLIP & 60.10{\tiny$\,\pm$3.37} & 56.28{\tiny$\,\pm$3.78} & 89.78{\tiny$\,\pm$1.18} & 69.56{\tiny$\,\pm$1.42} & 64.68{\tiny$\,\pm$1.38} & 95.13{\tiny$\,\pm$0.30} & 77.70{\tiny$\,\pm$2.14} & 77.77{\tiny$\,\pm$2.36} & 96.52{\tiny$\,\pm$0.44} \\
\midrule
\multicolumn{10}{l}{\textit{Fused sets (ours: OP alignment + combined graph + GCN)}} \\
KEEP$+$QuiltNet$+$PLIP & \textbf{79.63}{\tiny$\,\pm$2.58} & \textbf{77.40}{\tiny$\,\pm$2.95} & \underline{96.58}{\tiny$\,\pm$0.16} & \underline{91.45}{\tiny$\,\pm$1.32} & \underline{90.74}{\tiny$\,\pm$1.12} & \underline{99.67}{\tiny$\,\pm$0.08} & \textbf{90.84}{\tiny$\,\pm$0.99} & \textbf{90.87}{\tiny$\,\pm$1.11} & \textbf{99.10}{\tiny$\,\pm$0.18} \\
\quad $\Delta$ & \textcolor{darkblue}{\small $+1.09$} & \textcolor{darkblue}{\small $+1.78$} & \textcolor{darkblue}{\small $+0.90$} & \textcolor{darkblue}{\small $+2.12$} & \textcolor{darkblue}{\small $+2.84$} & \textcolor{darkblue}{\small $+0.49$} & \textcolor{darkblue}{\small $+4.04$} & \textcolor{darkblue}{\small $+4.04$} & \textcolor{darkblue}{\small $+0.72$} \\
\cmidrule(lr){1-10}
KEEP$+$PLIP$+$BiomedCLIP & \underline{79.39}{\tiny$\,\pm$2.44} & \underline{76.98}{\tiny$\,\pm$2.67} & \textbf{96.61}{\tiny$\,\pm$0.14} & \textbf{92.02}{\tiny$\,\pm$1.34} & \textbf{91.48}{\tiny$\,\pm$1.23} & \textbf{99.69}{\tiny$\,\pm$0.07} & \underline{90.53}{\tiny$\,\pm$0.83} & \underline{90.53}{\tiny$\,\pm$1.02} & \underline{99.03}{\tiny$\,\pm$0.11} \\
\quad $\Delta$ & \textcolor{darkblue}{\small $+0.85$} & \textcolor{darkblue}{\small $+1.36$} & \textcolor{darkblue}{\small $+0.93$} & \textcolor{darkblue}{\small $+2.69$} & \textcolor{darkblue}{\small $+3.58$} & \textcolor{darkblue}{\small $+0.51$} & \textcolor{darkblue}{\small $+3.73$} & \textcolor{darkblue}{\small $+3.7$} & \textcolor{darkblue}{\small $+0.65$} \\
\cmidrule(lr){1-10}
MUSK$+$QuiltNet$+$PLIP & 74.88{\tiny$\,\pm$2.50} & 72.21{\tiny$\,\pm$2.87} & 94.78{\tiny$\,\pm$0.15} & 86.40{\tiny$\,\pm$0.72} & 84.25{\tiny$\,\pm$0.34} & 99.12{\tiny$\,\pm$0.07} & 88.81{\tiny$\,\pm$0.79} & 88.82{\tiny$\,\pm$0.97} & 98.86{\tiny$\,\pm$0.19} \\
\quad $\Delta$ & \textcolor{darkblue}{\small $+4.54$} & \textcolor{darkblue}{\small $+4.61$} & \textcolor{darkblue}{\small $+2.04$} & \textcolor{darkblue}{\small $+3.21$} & \textcolor{darkblue}{\small $+4.39$} & \textcolor{darkblue}{\small $+0.52$} & \textcolor{darkblue}{\small $+3.65$} & \textcolor{darkblue}{\small $+3.65$} & \textcolor{darkblue}{\small $+0.5$} \\
\bottomrule
\end{tabular}}
\end{table}

\textbf{\dataset{LubLung}.}
\method{KQP} achieves the highest mean accuracy ($87.81\%$), macro-F1 ($83.48\%$), and AUC ($98.51\%$), improving upon \method{KEEP} by $5.16$, $5.95$, and $1.43$ percentage points, respectively. \method{KPB} also improves all three metrics over \method{KEEP}, although its macro-F1 ($82.36\%$) is $1.12$ percentage points below \method{KQP}. \method{MQP} produces the largest gains over its own constituents: $5.46$, $6.59$, and $1.84$ percentage points in accuracy, macro-F1, and AUC, respectively. Its final accuracy ($83.60\%$) and macro-F1 ($78.59\%$) exceed \method{KEEP} alone.

\begin{table}[H]
\centering
\caption{Accuracy, macro F1, and AUC (\%) at $K=8$ shots on \dataset{LubLung}, \dataset{BRACS}, \dataset{BACH}, for the vision-language backbones and the fused sets whose members are all vision-language models. Mean $\pm$ standard deviation over 5 random seeds. \textbf{Bold} = best, \underline{underlined} = second best in each column.}
\label{app:vlm_k8_b}
\setlength{\tabcolsep}{2.6pt}
\renewcommand{\arraystretch}{1.05}
\resizebox{\textwidth}{!}{%
\begin{tabular}{lccccccccc}
\toprule
Method & \multicolumn{3}{c}{LubLung} & \multicolumn{3}{c}{BRACS} & \multicolumn{3}{c}{BACH} \\
\cmidrule(lr){2-4}\cmidrule(lr){5-7}\cmidrule(lr){8-10}
& ACC & F1 & AUC & ACC & F1 & AUC & ACC & F1 & AUC \\
\midrule
\multicolumn{10}{l}{\textit{Vision-language foundation models (GCN adapter)}} \\
KEEP & 82.65{\tiny$\,\pm$0.99} & 77.53{\tiny$\,\pm$1.20} & 97.08{\tiny$\,\pm$0.55} & 53.07{\tiny$\,\pm$1.78} & 50.96{\tiny$\,\pm$2.34} & 84.74{\tiny$\,\pm$1.07} & 89.37{\tiny$\,\pm$3.15} & 89.35{\tiny$\,\pm$3.14} & 98.50{\tiny$\,\pm$0.02} \\
MUSK & 78.14{\tiny$\,\pm$2.29} & 72.00{\tiny$\,\pm$2.19} & 95.24{\tiny$\,\pm$0.72} & 46.71{\tiny$\,\pm$2.13} & 44.99{\tiny$\,\pm$0.49} & 81.68{\tiny$\,\pm$0.43} & 81.67{\tiny$\,\pm$4.11} & 81.47{\tiny$\,\pm$4.08} & 94.63{\tiny$\,\pm$1.62} \\
QuiltNet & 77.24{\tiny$\,\pm$0.86} & 71.76{\tiny$\,\pm$0.55} & 94.71{\tiny$\,\pm$0.19} & 44.43{\tiny$\,\pm$1.53} & 42.31{\tiny$\,\pm$2.24} & 76.32{\tiny$\,\pm$1.68} & 67.71{\tiny$\,\pm$5.09} & 67.82{\tiny$\,\pm$4.94} & 85.45{\tiny$\,\pm$4.05} \\
PLIP & 76.07{\tiny$\,\pm$1.33} & 70.84{\tiny$\,\pm$1.19} & 94.89{\tiny$\,\pm$0.24} & 42.24{\tiny$\,\pm$2.47} & 40.61{\tiny$\,\pm$1.85} & 76.51{\tiny$\,\pm$2.94} & 66.67{\tiny$\,\pm$9.59} & 66.91{\tiny$\,\pm$9.57} & 85.50{\tiny$\,\pm$4.00} \\
BiomedCLIP & 69.02{\tiny$\,\pm$1.16} & 63.30{\tiny$\,\pm$0.84} & 92.05{\tiny$\,\pm$0.68} & 43.46{\tiny$\,\pm$2.39} & 43.44{\tiny$\,\pm$2.33} & 78.76{\tiny$\,\pm$1.13} & 70.63{\tiny$\,\pm$1.73} & 70.52{\tiny$\,\pm$1.70} & 85.19{\tiny$\,\pm$0.41} \\
\midrule
\multicolumn{10}{l}{\textit{Fused sets (ours: OP alignment + combined graph + GCN)}} \\
KEEP$+$QuiltNet$+$PLIP & \textbf{87.81}{\tiny$\,\pm$1.41} & \textbf{83.48}{\tiny$\,\pm$0.83} & \textbf{98.51}{\tiny$\,\pm$0.07} & \textbf{55.83}{\tiny$\,\pm$2.48} & \textbf{53.23}{\tiny$\,\pm$3.31} & \textbf{86.71}{\tiny$\,\pm$0.34} & \underline{89.58}{\tiny$\,\pm$2.20} & \underline{89.53}{\tiny$\,\pm$2.29} & \underline{98.77}{\tiny$\,\pm$0.33} \\
\quad $\Delta$ & \textcolor{darkblue}{\small $+5.16$} & \textcolor{darkblue}{\small $+5.95$} & \textcolor{darkblue}{\small $+1.43$} & \textcolor{darkblue}{\small $+2.76$} & \textcolor{darkblue}{\small $+2.27$} & \textcolor{darkblue}{\small $+1.97$} & \textcolor{darkblue}{\small $+0.21$} & \textcolor{darkblue}{\small $+0.18$} & \textcolor{darkblue}{\small $+0.27$} \\
\cmidrule(lr){1-10}
KEEP$+$PLIP$+$BiomedCLIP & \underline{86.66}{\tiny$\,\pm$1.10} & \underline{82.36}{\tiny$\,\pm$1.30} & \underline{98.36}{\tiny$\,\pm$0.22} & \underline{54.74}{\tiny$\,\pm$2.00} & \underline{52.30}{\tiny$\,\pm$2.19} & \underline{86.35}{\tiny$\,\pm$0.36} & \textbf{89.79}{\tiny$\,\pm$2.68} & \textbf{89.78}{\tiny$\,\pm$2.77} & \textbf{98.89}{\tiny$\,\pm$0.24} \\
\quad $\Delta$ & \textcolor{darkblue}{\small $+4.01$} & \textcolor{darkblue}{\small $+4.83$} & \textcolor{darkblue}{\small $+1.28$} & \textcolor{darkblue}{\small $+1.67$} & \textcolor{darkblue}{\small $+1.34$} & \textcolor{darkblue}{\small $+1.61$} & \textcolor{darkblue}{\small $+0.42$} & \textcolor{darkblue}{\small $+0.43$} & \textcolor{darkblue}{\small $+0.39$} \\
\cmidrule(lr){1-10}
MUSK$+$QuiltNet$+$PLIP & 83.60{\tiny$\,\pm$1.17} & 78.59{\tiny$\,\pm$1.14} & 97.08{\tiny$\,\pm$0.78} & 50.53{\tiny$\,\pm$1.43} & 48.47{\tiny$\,\pm$1.16} & 83.84{\tiny$\,\pm$0.70} & 82.71{\tiny$\,\pm$4.11} & 82.60{\tiny$\,\pm$4.35} & 96.31{\tiny$\,\pm$1.20} \\
\quad $\Delta$ & \textcolor{darkblue}{\small $+5.46$} & \textcolor{darkblue}{\small $+6.59$} & \textcolor{darkblue}{\small $+1.84$} & \textcolor{darkblue}{\small $+3.82$} & \textcolor{darkblue}{\small $+3.48$} & \textcolor{darkblue}{\small $+2.16$} & \textcolor{darkblue}{\small $+1.04$} & \textcolor{darkblue}{\small $+1.13$} & \textcolor{darkblue}{\small $+1.68$} \\
\bottomrule
\end{tabular}}
\end{table}

\textbf{\dataset{BRACS}.}
\method{KQP} achieves the highest mean accuracy ($55.83\%$), macro-F1 ($53.23\%$), and AUC ($86.71\%$), with gains of $2.76$, $2.27$, and $1.97$ percentage points over \method{KEEP}. \method{KPB} follows with $54.74\%$ accuracy, $52.30\%$ macro-F1, and $86.35\%$ AUC. The ordering of these two configurations differs from $K=16$, where \method{KPB} leads across all three metrics. \method{MQP} achieves the largest gains over its constituents ($3.82$, $3.48$, and $2.16$ percentage points), but its final scores remain below \method{KEEP} alone. In particular, its macro-F1 increases from \method{MUSK}'s $44.99\%$ to $48.47\%$, compared with $50.96\%$ for \method{KEEP}.

\textbf{\dataset{BACH}.}
\method{KPB} achieves the highest mean accuracy ($89.79\%$), macro-F1 ($89.78\%$), and AUC ($98.89\%$), but its gains over \method{KEEP} are modest at $0.42$, $0.43$, and $0.39$ percentage points, respectively. \method{KQP} similarly yields small gains of $0.21$, $0.18$, and $0.27$ percentage points. The macro-F1 gains of these \method{KEEP}-based fusions are therefore substantially smaller than at $K=16$ ($2.86$-$3.06$ percentage points). \method{MQP} produces the largest gains over its own constituents across all three metrics, including $1.13$ percentage points in macro-F1 and $1.68$ in AUC.

Tables~\ref{app:vlm_k4_a} and~\ref{app:vlm_k4_b} report accuracy, macro-F1, and AUC at $K=4$ across the six datasets, complementing the main-text macro-F1 results in Table ~\ref{tab:f1_k4}.

\textbf{\dataset{LungHist700}.}
\method{KQP} achieves the highest mean accuracy ($70.63\%$) and macro-F1 ($66.76\%$), improving upon KEEP by $2.04$ and $1.43$ percentage points, respectively. \method{KPB} achieves the highest AUC ($94.61\%$), with a gain of $1.43$ percentage points over \method{KEEP}. \method{MQP} produces the largest gains over its constituents across all three metrics.

\textbf{\dataset{HeidelbergSkin}.}
\method{KPB} achieves the highest mean accuracy ($90.00\%$), macro-F1 ($88.38\%$), and AUC ($99.65\%$), with gains of $2.50$, $2.12$, and $0.61$ percentage points over \method{KEEP}. \method{KQP} follows closely, reaching $88.30\%$ macro-F1 and $99.61\%$ AUC. \method{MQP} achieves larger macro-F1 and AUC gains over its constituents ($3.38$ and $0.99$ percentage points), but its final scores remain below \method{KEEP} alone. The two \method{KEEP}-based fusions therefore retain the strongest absolute performance despite the larger relative improvements obtained by \method{MQP}.

\textbf{\dataset{Kather2016}.}
\method{MQP} achieves the highest mean accuracy ($82.94\%$), macro-F1 ($82.88\%$), and AUC ($97.85\%$), improving upon \method{MUSK} by $3.51$, $3.44$, and $0.77$ percentage points, respectively. The three fused configurations are closely grouped, with differences of at most $0.41$ percentage points in accuracy and macro-F1 and $0.22$ in AUC. \method{KQP} and \method{KPB} achieve larger macro-F1 gains of $6.07$ and $6.29$ percentage points. These gains are measured against \method{KEEP}'s lower baseline macro-F1 ($76.40\%$), whereas \method{MQP} is compared with MUSK ($79.44\%$). Thus, the larger gains of the \method{KEEP}-based fusions coexist with slightly lower final scores.

\begin{table}[H]
\centering
\caption{Accuracy, macro F1 and AUC (\%) at $K=4$ shots on \dataset{LungHist700}, \dataset{HeidelbergSkin}, \dataset{Kather2016}, for the vision-language backbones and the fused sets whose members are all vision-language models. Mean $\pm$ standard deviation over 5 random seeds. \textbf{Bold} = best, \underline{underlined} = second best in each column.}
\label{app:vlm_k4_a}
\setlength{\tabcolsep}{2.6pt}
\renewcommand{\arraystretch}{1.05}
\resizebox{\textwidth}{!}{%
\begin{tabular}{lccccccccc}
\toprule
Method & \multicolumn{3}{c}{LungHist700} & \multicolumn{3}{c}{HeidelbergSkin} & \multicolumn{3}{c}{Kather2016} \\
\cmidrule(lr){2-4}\cmidrule(lr){5-7}\cmidrule(lr){8-10}
& ACC & F1 & AUC & ACC & F1 & AUC & ACC & F1 & AUC \\
\midrule
\multicolumn{10}{l}{\textit{Vision-language foundation models (GCN adapter)}} \\
KEEP & 68.59{\tiny$\,\pm$3.54} & 65.33{\tiny$\,\pm$3.02} & 93.18{\tiny$\,\pm$0.92} & 87.50{\tiny$\,\pm$0.85} & 86.26{\tiny$\,\pm$0.83} & 99.04{\tiny$\,\pm$0.26} & 76.70{\tiny$\,\pm$1.18} & 76.40{\tiny$\,\pm$1.66} & 96.63{\tiny$\,\pm$0.78} \\
MUSK & 64.10{\tiny$\,\pm$3.92} & 60.52{\tiny$\,\pm$4.39} & 89.91{\tiny$\,\pm$1.61} & 78.97{\tiny$\,\pm$1.90} & 74.49{\tiny$\,\pm$2.75} & 97.62{\tiny$\,\pm$0.67} & 79.43{\tiny$\,\pm$3.27} & 79.44{\tiny$\,\pm$3.30} & 97.08{\tiny$\,\pm$1.22} \\
QuiltNet & 57.93{\tiny$\,\pm$1.02} & 53.15{\tiny$\,\pm$1.89} & 84.89{\tiny$\,\pm$3.78} & 69.30{\tiny$\,\pm$1.40} & 63.89{\tiny$\,\pm$1.27} & 94.92{\tiny$\,\pm$0.95} & 75.64{\tiny$\,\pm$3.65} & 75.42{\tiny$\,\pm$3.93} & 94.88{\tiny$\,\pm$1.24} \\
PLIP & 55.37{\tiny$\,\pm$3.97} & 50.94{\tiny$\,\pm$4.08} & 86.97{\tiny$\,\pm$1.68} & 69.67{\tiny$\,\pm$0.96} & 64.76{\tiny$\,\pm$0.57} & 96.28{\tiny$\,\pm$0.44} & 76.53{\tiny$\,\pm$2.68} & 76.34{\tiny$\,\pm$2.79} & 96.13{\tiny$\,\pm$0.82} \\
BiomedCLIP & 57.32{\tiny$\,\pm$6.56} & 52.62{\tiny$\,\pm$7.02} & 88.92{\tiny$\,\pm$2.45} & 62.74{\tiny$\,\pm$2.37} & 58.00{\tiny$\,\pm$3.83} & 93.81{\tiny$\,\pm$0.64} & 71.29{\tiny$\,\pm$4.31} & 70.72{\tiny$\,\pm$5.14} & 94.83{\tiny$\,\pm$1.38} \\
\midrule
\multicolumn{10}{l}{\textit{Fused sets (ours: OP alignment + combined graph + GCN)}} \\
KEEP$+$QuiltNet$+$PLIP & \textbf{70.63}{\tiny$\,\pm$6.52} & \textbf{66.76}{\tiny$\,\pm$6.63} & \underline{94.17}{\tiny$\,\pm$0.98} & \underline{89.58}{\tiny$\,\pm$1.16} & \underline{88.30}{\tiny$\,\pm$1.71} & \underline{99.61}{\tiny$\,\pm$0.08} & 82.53{\tiny$\,\pm$3.65} & 82.47{\tiny$\,\pm$3.62} & 97.63{\tiny$\,\pm$0.89} \\
\quad $\Delta$ & \textcolor{darkblue}{\small $+2.04$} & \textcolor{darkblue}{\small $+1.43$} & \textcolor{darkblue}{\small $+0.99$} & \textcolor{darkblue}{\small $+2.08$} & \textcolor{darkblue}{\small $+2.04$} & \textcolor{darkblue}{\small $+0.57$} & \textcolor{darkblue}{\small $+5.83$} & \textcolor{darkblue}{\small $+6.07$} & \textcolor{darkblue}{\small $+1.00$} \\
\cmidrule(lr){1-10}
KEEP$+$PLIP$+$BiomedCLIP & \underline{70.24}{\tiny$\,\pm$6.34} & \underline{66.19}{\tiny$\,\pm$6.87} & \textbf{94.61}{\tiny$\,\pm$0.79} & \textbf{90.00}{\tiny$\,\pm$1.47} & \textbf{88.38}{\tiny$\,\pm$2.23} & \textbf{99.65}{\tiny$\,\pm$0.05} & \underline{82.79}{\tiny$\,\pm$3.51} & \underline{82.69}{\tiny$\,\pm$3.48} & \underline{97.82}{\tiny$\,\pm$0.77} \\
\quad $\Delta$ & \textcolor{darkblue}{\small $+1.65$} & \textcolor{darkblue}{\small $+0.86$} & \textcolor{darkblue}{\small $+1.43$} & \textcolor{darkblue}{\small $+2.50$} & \textcolor{darkblue}{\small $+2.12$} & \textcolor{darkblue}{\small $+0.61$} & \textcolor{darkblue}{\small $+6.09$} & \textcolor{darkblue}{\small $+6.29$} & \textcolor{darkblue}{\small $+1.19$} \\
\cmidrule(lr){1-10}
MUSK$+$QuiltNet$+$PLIP & 66.34{\tiny$\,\pm$3.75} & 62.08{\tiny$\,\pm$4.37} & 92.50{\tiny$\,\pm$1.14} & 81.40{\tiny$\,\pm$1.21} & 77.87{\tiny$\,\pm$1.24} & 98.61{\tiny$\,\pm$0.11} & \textbf{82.94}{\tiny$\,\pm$4.29} & \textbf{82.88}{\tiny$\,\pm$4.36} & \textbf{97.85}{\tiny$\,\pm$0.93} \\
\quad $\Delta$ & \textcolor{darkblue}{\small $+2.24$} & \textcolor{darkblue}{\small $+1.56$} & \textcolor{darkblue}{\small $+2.59$} & \textcolor{darkblue}{\small $+2.43$} & \textcolor{darkblue}{\small $+3.38$} & \textcolor{darkblue}{\small $+0.99$} & \textcolor{darkblue}{\small $+3.51$} & \textcolor{darkblue}{\small $+3.44$} & \textcolor{darkblue}{\small $+0.77$} \\
\bottomrule
\end{tabular}}
\end{table}

\textbf{\dataset{LubLung}.}
\method{KQP} achieves the highest mean accuracy ($82.40\%$), macro-F1 ($76.08\%$), and AUC ($97.45\%$), with gains of $3.05$, $3.98$, and $1.82$ percentage points over \method{KEEP}. \method{KPB} performs similarly, differing by at most $0.26$ percentage points across the three metrics. \method{MQP} yields the largest gains over its constituents: $6.49$, $6.54$, and $2.39$ percentage points in accuracy, macro-F1, and AUC, respectively. For \method{MQP}, the strongest constituent is \method{QuiltNet} for accuracy and macro-F1, but \method{MUSK} for AUC. Despite these larger gains, \method{MQP}'s final macro-F1 ($68.98\%$) remains below \method{KEEP} alone ($72.10\%$).

\textbf{\dataset{BRACS}.}
\method{KPB} achieves the highest mean accuracy ($50.95\%$), macro-F1 ($49.53\%$), and AUC ($85.64\%$), improving upon \method{KEEP} by $1.48$, $0.89$, and $0.72$ percentage points, respectively. \method{KQP} improves accuracy and AUC by $0.53$ percentage points each, but its macro-F1 decreases from $48.64\%$ to $48.33\%$. \method{MQP} improves all three metrics over MUSK, including a macro-F1 increase of $3.55$ percentage points. Its final macro-F1 ($42.22\%$), however, remains below \method{KEEP} and both \method{KEEP}-based fusions.

\begin{table}[H]
\centering
\caption{Accuracy, macro F1, and AUC (\%) at $K=4$ shots on \dataset{LubLung}, \dataset{BRACS}, \dataset{BACH}, for the vision-language backbones and the fused sets whose members are all vision-language models. Mean $\pm$ standard deviation over 5 random seeds. \textbf{Bold} = best, \underline{underlined} = second best in each column.}
\label{app:vlm_k4_b}
\setlength{\tabcolsep}{2.6pt}
\renewcommand{\arraystretch}{1.05}
\resizebox{\textwidth}{!}{%
\begin{tabular}{lccccccccc}
\toprule
Method & \multicolumn{3}{c}{LubLung} & \multicolumn{3}{c}{BRACS} & \multicolumn{3}{c}{BACH} \\
\cmidrule(lr){2-4}\cmidrule(lr){5-7}\cmidrule(lr){8-10}
& ACC & F1 & AUC & ACC & F1 & AUC & ACC & F1 & AUC \\
\midrule
\multicolumn{10}{l}{\textit{Vision-language foundation models (GCN adapter)}} \\
KEEP & 79.35{\tiny$\,\pm$3.12} & 72.10{\tiny$\,\pm$4.99} & 95.63{\tiny$\,\pm$1.79} & 49.47{\tiny$\,\pm$5.62} & \underline{48.64}{\tiny$\,\pm$4.91} & \underline{84.92}{\tiny$\,\pm$0.57} & \textbf{86.46}{\tiny$\,\pm$9.73} & \textbf{86.26}{\tiny$\,\pm$9.94} & \underline{97.59}{\tiny$\,\pm$1.54} \\
MUSK & 68.80{\tiny$\,\pm$2.79} & 61.51{\tiny$\,\pm$3.18} & 92.73{\tiny$\,\pm$1.20} & 40.44{\tiny$\,\pm$3.58} & 38.67{\tiny$\,\pm$2.09} & 79.54{\tiny$\,\pm$1.10} & 75.42{\tiny$\,\pm$6.99} & 75.66{\tiny$\,\pm$6.89} & 92.22{\tiny$\,\pm$2.61} \\
QuiltNet & 69.15{\tiny$\,\pm$1.62} & 62.44{\tiny$\,\pm$1.68} & 91.68{\tiny$\,\pm$0.27} & 36.49{\tiny$\,\pm$2.17} & 34.83{\tiny$\,\pm$2.90} & 73.45{\tiny$\,\pm$3.21} & 52.92{\tiny$\,\pm$5.67} & 52.17{\tiny$\,\pm$6.43} & 78.74{\tiny$\,\pm$2.26} \\
PLIP & 66.75{\tiny$\,\pm$3.59} & 61.50{\tiny$\,\pm$1.50} & 92.04{\tiny$\,\pm$0.70} & 33.64{\tiny$\,\pm$3.01} & 32.35{\tiny$\,\pm$2.63} & 73.34{\tiny$\,\pm$1.22} & 56.25{\tiny$\,\pm$6.61} & 55.72{\tiny$\,\pm$7.00} & 78.51{\tiny$\,\pm$2.36} \\
BiomedCLIP & 63.17{\tiny$\,\pm$2.03} & 55.96{\tiny$\,\pm$0.75} & 90.48{\tiny$\,\pm$1.21} & 38.99{\tiny$\,\pm$3.12} & 37.65{\tiny$\,\pm$2.81} & 77.50{\tiny$\,\pm$1.08} & 62.71{\tiny$\,\pm$7.92} & 62.04{\tiny$\,\pm$8.48} & 83.05{\tiny$\,\pm$3.09} \\
\midrule
\multicolumn{10}{l}{\textit{Fused sets (ours: OP alignment + combined graph + GCN)}} \\
KEEP$+$QuiltNet$+$PLIP & \textbf{82.40}{\tiny$\,\pm$3.21} & \textbf{76.08}{\tiny$\,\pm$4.62} & \textbf{97.45}{\tiny$\,\pm$0.43} & \underline{50.00}{\tiny$\,\pm$3.53} & 48.33{\tiny$\,\pm$2.32} & 85.45{\tiny$\,\pm$0.85} & 84.83{\tiny$\,\pm$3.43} & 84.61{\tiny$\,\pm$3.57} & 97.33{\tiny$\,\pm$1.30} \\
\quad $\Delta$ & \textcolor{darkblue}{\small $+3.05$} & \textcolor{darkblue}{\small $+3.98$} & \textcolor{darkblue}{\small $+1.82$} & \textcolor{darkblue}{\small $+0.53$} & \textcolor{gainneg}{\small $-0.31$} & \textcolor{darkblue}{\small $+0.53$} & \textcolor{gainneg}{\small $-1.63$} & \textcolor{gainneg}{\small $-1.65$} & \textcolor{gainneg}{\small $-0.26$} \\
\cmidrule(lr){1-10}
KEEP$+$PLIP$+$BiomedCLIP & \underline{82.14}{\tiny$\,\pm$2.02} & \underline{76.02}{\tiny$\,\pm$3.67} & \underline{97.40}{\tiny$\,\pm$0.72} & \textbf{50.95}{\tiny$\,\pm$4.00} & \textbf{49.53}{\tiny$\,\pm$2.69} & \textbf{85.64}{\tiny$\,\pm$0.97} & \underline{86.17}{\tiny$\,\pm$2.64} & \underline{85.94}{\tiny$\,\pm$2.78} & \textbf{97.61}{\tiny$\,\pm$1.24} \\
\quad $\Delta$ & \textcolor{darkblue}{\small $2.79$} & \textcolor{darkblue}{\small $+3.92$} & \textcolor{darkblue}{\small $+1.77$} & \textcolor{darkblue}{\small $+1.48$} & \textcolor{darkblue}{\small $+0.89$} & \textcolor{darkblue}{\small $+0.72$} & \textcolor{gainneg}{\small $-0.29$} & \textcolor{gainneg}{\small $-0.32$} & \textcolor{darkblue}{\small $+0.02$} \\
\cmidrule(lr){1-10}
MUSK$+$QuiltNet$+$PLIP & 75.64{\tiny$\,\pm$1.22} & 68.98{\tiny$\,\pm$2.28} & 95.12{\tiny$\,\pm$0.83} & 44.28{\tiny$\,\pm$4.20} & 42.22{\tiny$\,\pm$3.48} & 81.54{\tiny$\,\pm$0.69} & 74.67{\tiny$\,\pm$5.29} & 74.23{\tiny$\,\pm$5.80} & 92.80{\tiny$\,\pm$3.45} \\
\quad $\Delta$ & \textcolor{darkblue}{\small $+6.49$} & \textcolor{darkblue}{\small $+6.54$} & \textcolor{darkblue}{\small $+2.39$} & \textcolor{darkblue}{\small $+3.84$} & \textcolor{darkblue}{\small $+3.55$} & \textcolor{darkblue}{\small $+2.00$} & \textcolor{gainneg}{\small $-0.75$} & \textcolor{gainneg}{\small $-1.43$} & \textcolor{darkblue}{\small $+0.58$} \\
\bottomrule
\end{tabular}}
\end{table}

\textbf{\dataset{BACH}.}
\method{KEEP} retains the highest mean accuracy ($86.46\%$) and macro-F1 ($86.26\%$). All three fused sets yield lower accuracy and macro-F1 than their strongest constituent, with macro-F1 decreases of $1.65$, $0.32$, and $1.43$ percentage points for \method{KQP}, \method{KPB}, and \method{MQP}, respectively. AUC follows a different pattern: \method{KPB} achieves the highest value ($97.61\%$), although its gain over \method{KEEP} is only $0.02$ percentage points. \method{MQP} also improves AUC by $0.58$ percentage points despite its lower accuracy and macro-F1. For the \method{KEEP}-based fusions, macro-F1 standard deviations decrease from $9.94$ for \method{KEEP} to $3.57$ for \method{KQP} and $2.78$ for \method{KPB}.

\end{document}